\documentclass[nohyperref]{article}

\usepackage{microtype}
\usepackage{graphicx}
\usepackage{subfigure}
\usepackage{booktabs} %
\usepackage{makecell}

\usepackage{hyperref}

\usepackage[accepted]{icml2022}

\usepackage{amsmath}
\usepackage{amssymb}
\usepackage{mathtools}
\usepackage{amsthm}
\usepackage{chngcntr}

\usepackage{nicefrac}
\usepackage{bm}
\usepackage{booktabs}
\usepackage{amsfonts}
\usepackage{bbm}
\usepackage{mleftright}
\usepackage{dsfont}
\usepackage{placeins}
\usepackage{amscd,amssymb,amsmath,amsthm,bbold}
\usepackage[normalem]{ulem}
\usepackage{graphicx}
\usepackage{float}
\usepackage{multirow}
\usepackage{wrapfig}
\usepackage{varwidth}
\usepackage{xcolor}

\usepackage{nicefrac}
\usepackage{bm}
\usepackage{booktabs}
\usepackage{amsfonts}
\usepackage{bbm}
\usepackage{mleftright}
\usepackage{dsfont}
\usepackage{placeins}
\usepackage{amscd,amssymb,amsmath,amsthm,bbold}
\usepackage[normalem]{ulem}
\usepackage[font=small]{caption}

\makeatletter
\renewcommand{\ALG@name}{Recipe}
\makeatother

\DeclareMathOperator*{\argmin}{arg\,min}
\DeclareMathOperator*{\argmax}{arg\,max}

\definecolor{limegreen}{rgb}{0.2, 0.8, 0.2}

\usepackage[capitalize,noabbrev]{cleveref}

\theoremstyle{plain}

\theoremstyle{definition}

\theoremstyle{remark}

\mathchardef\mhyphen="2D

\usepackage[disable,textsize=tiny]{todonotes}
\newcommand{\yair}[1]{\todo{Yair: #1}}
\newcommand{\hong}[1]{\todo{Hong: #1}}
\newcommand{\rapha}[1]{\todo{Rapha: #1}}

\icmltitlerunning{Model soups: averaging weights of multiple fine-tuned
models improves accuracy without increasing inference time
}

\begin{document}

\twocolumn[
\icmltitle{Model soups: averaging weights of multiple fine-tuned
models

improves accuracy without increasing inference time}

\icmlsetsymbol{equal}{*}
\begin{icmlauthorlist}
\icmlauthor{Mitchell Wortsman}{uw}
\icmlauthor{Gabriel Ilharco}{uw}
\icmlauthor{Samir Yitzhak Gadre}{col}
\icmlauthor{Rebecca Roelofs}{goo}
\icmlauthor{Raphael Gontijo-Lopes}{goo}
\icmlauthor{Ari S. Morcos}{fb}
\icmlauthor{Hongseok Namkoong}{col}
\icmlauthor{Ali Farhadi}{uw}
\icmlauthor{Yair Carmon}{equal,tel}
\icmlauthor{Simon Kornblith}{equal,goo}
\icmlauthor{Ludwig Schmidt}{equal,uw}
\end{icmlauthorlist}

\icmlaffiliation{uw}{University of Washington}
\icmlaffiliation{col}{Columbia University}
\icmlaffiliation{goo}{Google Research, Brain Team}
\icmlaffiliation{tel}{Tel Aviv University}
\icmlaffiliation{fb}{Meta AI Research}

\icmlcorrespondingauthor{}{mitchnw@uw.edu}

\icmlkeywords{fine-tuning, weight-averaging, robustness, zero-shot, pre-training}

\vskip 0.3in
]

\printAffiliationsAndNotice{\icmlEqualContribution} %

\begin{abstract}
    The conventional recipe for maximizing model accuracy is to (1) train multiple models with various hyperparameters and (2) pick the individual model which performs best on a held-out validation set, discarding the remainder.
    In this paper, we revisit the second step of this procedure in the context of fine-tuning large pre-trained models, 
    where fine-tuned models often appear to lie in a single low error basin.
    We show that averaging the weights of multiple models fine-tuned with different hyperparameter configurations often improves accuracy and robustness. 
    Unlike a conventional ensemble, we may average many models without incurring any additional inference or memory costs---we call the results ``model soups.''
    When fine-tuning large pre-trained models such as CLIP, ALIGN, and a ViT-G pre-trained on JFT, our soup recipe provides significant improvements over the best model in a hyperparameter sweep on ImageNet.
    The resulting ViT-G model, which attains 90.94\% top-1 accuracy on ImageNet, achieved a new state of the art.
    Furthermore, we show that the model soup approach extends to multiple image classification and natural language processing tasks, improves out-of-distribution performance, and improves zero-shot performance on new downstream tasks.
    Finally, we analytically relate the performance similarity of weight-averaging and logit-ensembling to flatness of the loss and confidence of the predictions, and validate this relation empirically.
    Code is available at {\small\url{https://github.com/mlfoundations/model-soups}}.
\end{abstract}       
        \begin{figure}
            \centering
            \includegraphics[width=1\columnwidth]{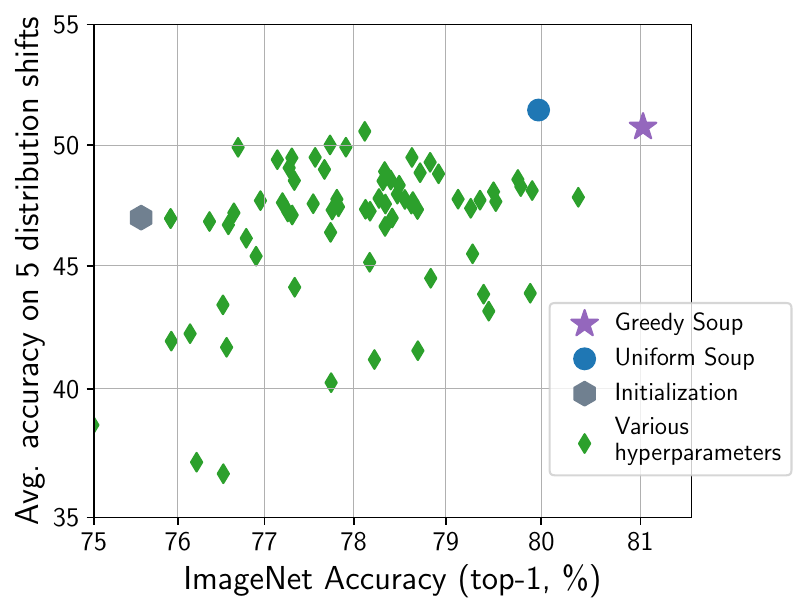}
            \vspace*{-0.7cm}
            \captionof{figure}{
                \emph{Model soups} improve accuracy over the best individual model when
                performing a large, random hyperparameter search
                for fine-tuning a CLIP ViT-B/32
                model on ImageNet. 
                The \textit{uniform soup} (blue circle) averages all fine-tuned
                models (green diamonds) in a random hyperparameter search over learning rate, weight-decay, iterations, data augmentation, mixup, and label smoothing.
                The \emph{greedy soup} adds models sequentially to the model soup, keeping a model in the soup
                if accuracy on the held-out validation set does not decrease.
            }
            \label{fig:teaser}
        \end{figure}
            \begin{table}[h!]
                \footnotesize
                \setlength{\tabcolsep}{5pt}
                \begin{tabular}{ l l l l }
                  \toprule
                  Method & ImageNet acc. & Distribution\\
                  &  (top-1, \%) & shifts\\ 
                  \midrule
                  ViT-G~\cite{zhai2021scaling} & 90.45 & -- \\
                  CoAtNet-7~\cite{dai2021coatnet} & 90.88 & -- \\
                  \midrule
                  \multicolumn{3}{l}{\textit{\color{gray}Our models/evaluations based on ViT-G:}}\\
                  ViT-G (reevaluated) & 90.47 & 82.06 \\
                  Best model in & 90.78 & 84.68 \\
                  hyperparam search &  &  \\
            
                  Greedy soup & \textbf{90.94} & \textbf{85.02} \\
            
                  \bottomrule
                \end{tabular}
                \vspace*{-0.05cm}
                \caption{
                \emph{Model soups} improve accuracy over the best individual model when fine-tuning a JFT-3B pre-trained
                ViT-G/14 model on ImageNet.
                Instead of selecting the best model from a hyperparameter sweep during fine-tuning, \emph{model soups} average the weights of multiple fine-tuned models.
                To evaluate performance under distribution shift we consider average accuracy on ImageNet-V2, ImageNet-R,
                ImageNet-Sketch,
                ObjectNet, and ImageNet-A.
                Additional details are provided
                by Table~\ref{tab:vit_g_finetuned_result} and
                Section~\ref{sec:jft}, while analogous results for BASIC~\cite{pham2021scaling} are in Appendix~\ref{app:basic}.}
                \label{tab:jft}
            \end{table}
            \FloatBarrier
            \section{Introduction} 
            In recent years, research has shown that models pre-trained on large and diverse datasets learn representations that transfer well to a variety of tasks. 
            As a result, machine learning practitioners now commonly develop solutions for downstream tasks by fine-tuning large pre-trained models \cite{girshick2014rich,yosinski2014transferable,kornblith2019better,kolesnikov2020big}. 
            Typically, the fine-tuning process involves two steps:
            (1) fine-tune models with a variety of hyperparameter configurations, and  
            (2) select the model which achieves the highest accuracy on the held-out validation set.
            The remaining models are then discarded.

        Selecting a single model and discarding the rest has several downsides.  For one, ensembling outputs of many models
        can outperform the best single model,
        albeit at a high computational cost during inference.
        \hong{In addition to Yair's comment, I think we should emphasize that inference cost is a BIG DEAL. Some researchers have the impression that training cost is the only elephant in the room. But the computational multiplier in front inference cost is often at least in the order of 10Ms to Bs and inference cost is often a bigger cost overall.}
        For another, fine-tuning a model on downstream tasks can sometimes reduce out-of-distribution performance \cite{radford2021learning,andreassen2021evolution,wortsman2021robust,pham2021scaling}, %
        and the best single model on the target distribution may not be the best model on out-of-distribution data.
        
        In this work, we propose a more accurate and robust alternative to the second step of the conventional recipe in the context of fine-tuning a large pre-trained model. Instead of selecting the individual fine-tuned model which achieves the highest accuracy on the held-out validation set, we average the weights of models fine-tuned independently, and refer to the result as a \emph{model soup}. Given the results of the first step---a hyperparameter sweep over fine-tuned models---averaging several of these models to form a model soup requires no additional training and adds no cost at inference time.

        Since the loss landscape of neural network training is non-convex with many solutions in different loss basins, it is perhaps surprising that averaging the weights of independently fine-tuned models achieves high performance.  However, recent work \cite{neyshabur2020being} observes that fine-tuned models optimized independently from the same pre-trained initialization lie in the same basin of the error landscape, inspiring our method.
        Weight averaging along a single training trajectory has previously been shown to improve the performance of models in non-transfer settings~\cite{szegedy2016rethinking,izmailov2018averaging}. Our approach extends weight averaging to the context of fine-tuning, where we find that it also works across many independent runs with varied hyperparemeter configurations.
        Our use of a diverse set of fine-tuned models is inspired by~\citet{gontijo2021no} who observe that ensembling independent runs trained with different hyperparameters improves performance.
        
        \rapha{In addition to Hong's comment about outline (below), it might also be useful to add a bold "Contributions" in this paragraph, this could replace a bulletpoint contributions list if you think the detailed contributions paragraph is more informative.}
        We perform a comprehensive experimental study of fine-tuning to understand the behavior of model soups. 
        For our main results we fine-tune CLIP~\cite{radford2021learning} and ALIGN~\cite{jia2021scaling}, which are pre-trained with a contrastive loss on image-text pairs, and a ViT-G model pre-trained on JFT~\cite{zhai2021scaling}. 
        Our results show that model soups often outperform the best individual model on both the in-distribution and natural distribution shift test sets (Table~\ref{tab:jft}, Figure~\ref{fig:teaser}, Figure~\ref{fig:align}).
        A model soup composed of ViT-G models achieves 90.94\% on ImageNet~\cite{deng2009imagenet}, surpassing the previous state of the art of 90.88\% attained by the CoAtNet model~\cite{dai2021coatnet} while requiring 25\% fewer FLOPs at inference time.\footnote{Since our initial submission, we attain 90.98\% with BASIC~\cite{pham2021scaling}, which ties the newer CoCa model~\cite{yu2022coca} to their reported precision; see Appendix~\ref{app:basic}.}
        In general, model soups can approach the performance of ensembling,
        with no additional computational cost or memory relative to a single model during inference.
        Beyond ImageNet and associated distribution shifts, our results show that model soups are applicable
        when fine-tuning on tasks from the WILDS~\cite{wilds2021} benchmark, and when fine-tuning transformer models~\cite{vaswani2017attention,devlin-etal-2019-bert,raffel2020t5} for text classification.

        While the most straightforward approach to making a model soup is to average all the weights uniformly, we find that \emph{greedy soups}, where models are sequentially added to the soup if they improve accuracy on held-out data, outperforms uniform averaging. Greedy soups avoid adding in models which may lie in a different basin of the error landscape, which could happen if, for example, models are fine-tuned with high learning rates.
        
        In addition to empirical observations, we analytically relate the similarity in loss between weight-averaging and logit-ensembling to the flatness of the loss (i.e., its second derivative on a line between models) and confidence of the predictions (expressed via the variance of a logits difference drawn from the weight-average softmax). We empirically validate our approximation on a subset of the models we train and show that it is strongly correlated with the true averaging vs.\ ensembling performance difference, particularly in the learning rate regimes where soups are effective and models achieve higher accuracy.
        
        \textbf{Paper outline.} Our method of \emph{model soups} is presented and evaluated in Sections \ref{sec:method} and \ref{sec:exp}, respectively.
        Next, Section~\ref{sec:theory} includes our analysis relating model soups and ensembles, Section~\ref{sec:lim} details the scope and limitations of the proposed method, and Section~\ref{sec:related} contextualizes \emph{model soups} by reviewing related work.

        \section{Method} \label{sec:method}
        \begin{table}
        \renewcommand{\arraystretch}{1.15}
            \caption{The primary methods contrasted in this work.
            Each $\theta_i$ is a model found through fine-tuning from 
            a shared initialization.
            Cost refers to the memory and compute requirements during inference
            relative to a single model. 
            All methods require the same training.}
            \vspace*{-0.2cm}
            \begin{center}
                \begin{tabular}{lcc} 
                \toprule
                 & Method & Cost \\
                 \midrule
                 Best on val. set & $f\mleft(x, \argmax_{i} \mathsf{ValAcc}\mleft(\theta_i\mright) \mright)$  & $\mathcal{O}(1)$  \\
                 Ensemble & $\frac{1}{k}\sum_{i=1}^k f\mleft(x, \theta_i\mright)$ & $\mathcal{O}(k)$ \\ %
                 Uniform soup & $f\mleft(x,  \frac{1}{k}\sum_{i=1}^k \theta_i\mright)$ & $\mathcal{O}(1)$  \\
                 Greedy soup & Recipe~\ref{alg:greedy} & $\mathcal{O}(1)$  \\ 
                 Learned soup & Appendix~\ref{app:learned-soup} & $\mathcal{O}(1)$  \\  
                 \bottomrule
                \end{tabular}
            \end{center}
            \label{tab:methods}
            \vspace*{-0.4cm}
        \end{table}

        This section highlights three recipes for model souping, the \emph{uniform}, \emph{greedy},
        and \emph{learned} soup, though the greedy soup is our central method.
        We summarize the methods described in this section in Table~\ref{tab:methods}.
        
        We consider a neural network $f(x, \theta)$ with input data $x$ and parameters
        $\theta \in \mathbb{R}^d$. Fine-tuning is analogous
        to standard neural network training but includes an important distinction: the parameters
        are initialized to those found via pre-training.
        
        Let $\theta = \mathsf{FineTune}\mleft(\theta_0, h\mright)$ denote the
        parameters obtained by fine-tuning with pre-trained initialization $\theta_0$ and
        hyperparameter configuration $h$.
        The hyperparameter configuration can include the
        choice of optimizer, data augmentation, training iterations, and a random
        seed which will determine data order. 
        
        For hyperparameter
        configurations $h_1,...,h_k$ let
        $\theta_i = \mathsf{FineTune}\mleft(\theta_0, h_i\mright)$.
        Conventionally, the parameters $\theta_j$ which attain the highest
        accuracy on a held out validation set  are selected, and the remaining parameters are discarded.
        Instead, \emph{model soups} $f\mleft(x, \theta_{\mathcal{S}}\mright)$ use an average
        of $\theta_i$, i.e., $ \theta_{\mathcal{S}} = \frac{1}{|\mathcal{S}|}\sum_{i \in \mathcal{S}} \theta_i$
        where $\mathcal{S} \subseteq \{1,...,k\}$.
        The \emph{uniform soup}
        is constructed by averaging \emph{all} fine-tuned models $\theta_i$ and so $\mathcal{S} = \{1,...,n\}$.

        There are settings in which 
        a hyperparameter configuration can produce
        a model with low accuracy that results in a low accuracy uniform soup.
        This issue can be circumvented with a \emph{greedy soup} (Recipe~\ref{alg:greedy}).
        The greedy soup is constructed by sequentially adding each model as
        a potential ingredient in the soup, and only keeping the model in the soup
        if performance on a held out validation set (disjoint from the training and test sets) improves.
        Before running this procedure we
        sort the models in decreasing order of validation set accuracy, and so the greedy soup can be
        no worse than the best individual model on the held-out validation set. We also explore a more advanced \emph{learned soup} recipe that optimizes model interpolation weights by gradient-based minibatch optimization (see Appendix~\ref{app:learned-soup} for details). This procedure requires simultaneously loading all models in memory which currently 
        hinders its use with large networks.
    
         \begin{algorithm}[tb]
            \caption{$\mathsf{GreedySoup}$}
            \label{alg:greedy}
         \begin{algorithmic}
        \STATE {\bfseries Input:} Potential soup ingredients $\{\theta_1, ... , \theta_k \}$ (sorted in decreasing order of $\mathsf{ValAcc}\mleft(\theta_i\mright)$).
        \STATE $\mathsf{ingredients} \gets \{ \}$
        \STATE {\bfseries for} $i = 1$ {\bfseries to} $k$ {\bfseries do}
        \STATE \ \ \ {\bfseries if}  $\mathsf{ValAcc}\mleft(\mathsf{average}\mleft(\mathsf{ingredients} \cup \{ \theta_i \}\mright)\mright)\geq$
        \STATE \ \ \ \ \ \ \ \ \ \ \ \ \ \ \ \ \ \ $\mathsf{ValAcc}\mleft(\mathsf{average}\mleft(\mathsf{ingredients}\mright)\mright)$ {\bfseries then}
        \STATE \ \ \ \ \ \ \ $\mathsf{ingredients} \gets \mathsf{ingredients} \cup \{\theta_i \}$
        \STATE {\bfseries return} $\mathsf{average}\mleft(\mathsf{ingredients}\mright)$
         \end{algorithmic}
         \end{algorithm}
         
        \begin{figure*}
            \centering
            \vspace*{-0.2cm}
            \includegraphics[width=\textwidth]{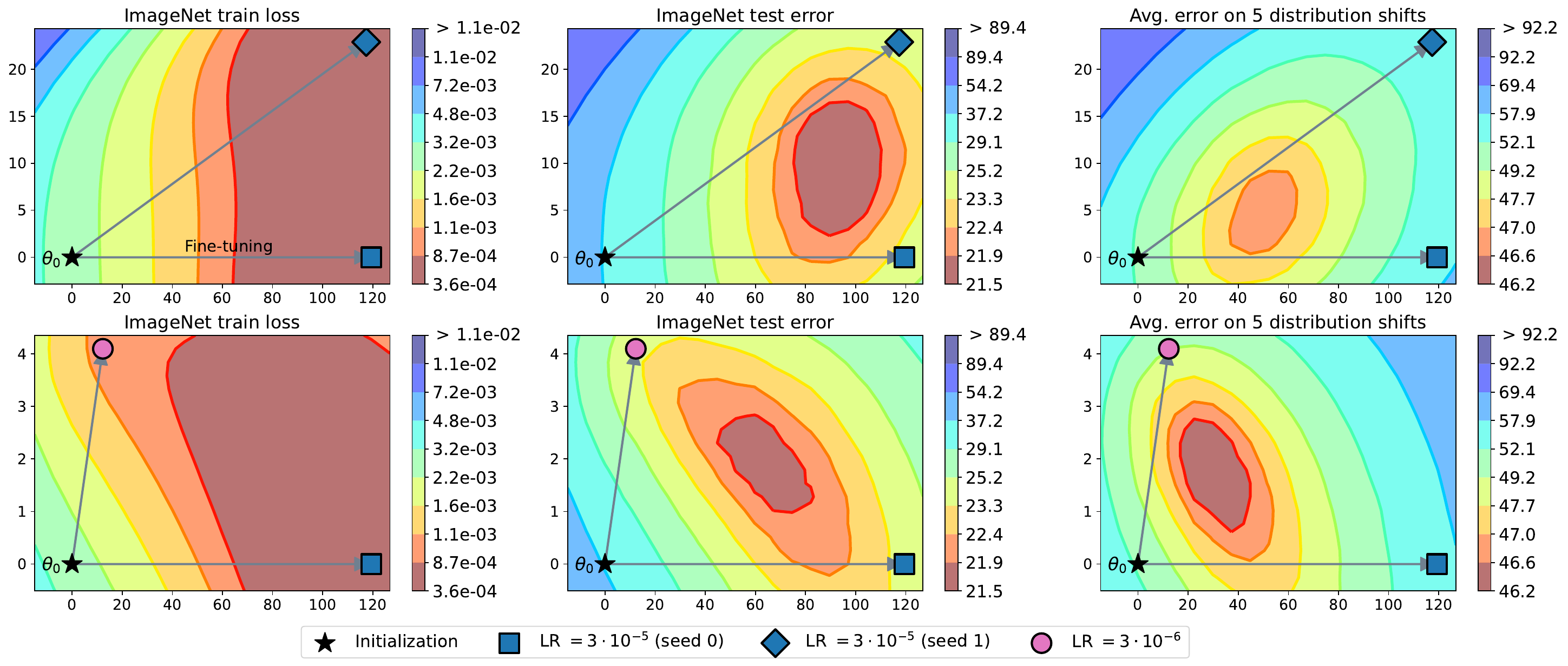}
            \caption{
            The solution with the highest accuracy is often not a fine-tuned model but rather lies between fine-tuned models.
            This figure shows loss and error on a two dimensional slice of the loss and error landscapes.
            We use the zero-shot initialization $\theta_0$ and fine-tune twice (illustrated by the gray arrows), independently, to obtain solutions $\theta_1$
            and $\theta_2$. As in~\citet{garipov2018loss}, we obtain an orthonormal basis $u_1$, $u_2$ for the plane
            spanned by these models, and the $x$ and $y$-axis show movement in parameter space in these directions, respectively.
            }
            \label{fig:error}
            \vspace*{-0.3cm}
        \end{figure*}
        \section{Experiments}\label{sec:exp}
        This section presents our key experimental findings.
        We begin with experimental setup (Section~\ref{sec:setup})
        then provide intuition for model soups by examining error landscape
        visualizations (Section~\ref{sec:error}).
        Next we present our main results (Section~\ref{sec:mainres}), using model soups 
        as an alternative to selecting the best performing individual model.
        The appendix includes additional results on model soups in the context of robust fine-tuning (Appendix~\ref{sec:robustft})
        and model soups constructed by fine-tuning on different datasets (Appendix~\ref{sec:zsperf}).
        
        \subsection{Experimental setup}\label{sec:setup}
        
        Our experiments explore the application of model soups when fine-tuning various models.
        The primary models we fine-tune are
        the CLIP \cite{radford2021learning}, ALIGN \cite{jia2021scaling}, and BASIC~\cite{pham2021scaling} 
        models pre-trained with contrastive supervision from image-text pairs,
        a ViT-G/14 model pre-trained on JFT-3B~\cite{zhai2021scaling},
        and transformer models for text classification \cite{devlin-etal-2019-bert,colin2020exploring}.
        Unless otherwise mentioned, experiments use the CLIP ViT-B/32 model.
        Fine-tuning is performed end-to-end (all parameters are modified)
        which typically results
        in better accuracy than training only the final linear layer \cite{kornblith2019better,agrawal2014analyzing,chatfield2014return,azizpour2015generic}.
        
        We consider two different methods for initializing the final linear layer before fine-tuning.
        The first method initializes the model from a linear probe (LP), as described in~\citet{kumar2021finetuning},
        and we refer to this method as LP initialization.
        The second method uses the zero-shot initialization, e.g., using the classifier produced by the text tower of CLIP or ALIGN as the initialization.
        Both methods for initializing the model produce similar trends when applicable, and unless otherwise
        stated we use the LP initialization.

        For the ensemble baselines~\cite{dietterich2000ensemble,deepensembles} we ensemble the logits (unormalized outputs)
        of models as in~\citet{gontijo2021no}.
        Fine-tuning uses a supervised cross-entropy loss and, unless otherwise mentioned, 
        is conducted on ImageNet~\cite{deng2009imagenet}.
        When fine-tuning on ImageNet we also evaluate on the five natural distribution shifts:
        ImageNetV2~\cite{pmlr-v97-recht19a}, ImageNet-R~\cite{imagenetr}, ImageNet-Sketch~\cite{imagenetsketch},
        ObjectNet~\cite{objectnet}, and ImageNet-A~\cite{imageneta}.
        We often report results averaged over these five distribution shifts.
        Since the official ImageNet validation set is typically used as the test set,
        we use roughly 2\% of the ImageNet training set as a held-out validation
        set for constructing greedy soups.
        
        \subsection{Intuition and motivation}\label{sec:error}
            \begin{figure}
            \centering
            \includegraphics[width=0.9\columnwidth]{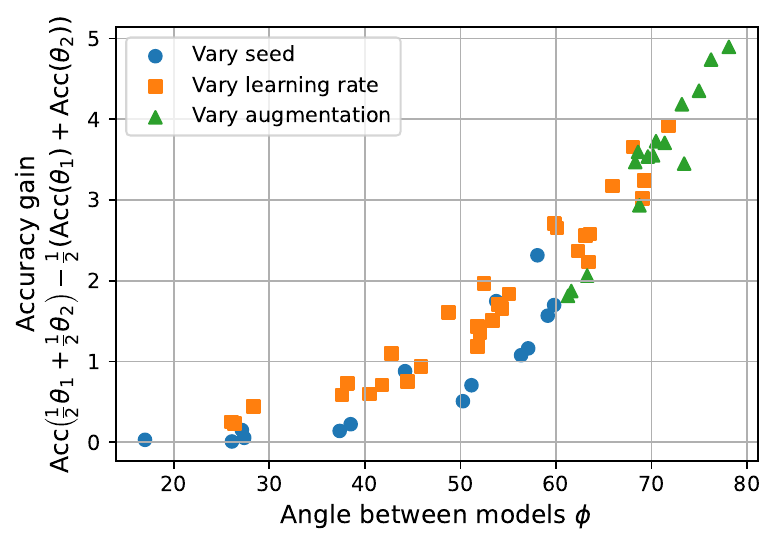}
            \vspace*{-0.2cm}
            \caption{
                The advantage of averaging solutions ($y$-axis) is correlated with the angle $\phi$ between between solutions,
                while varying hyperparameter configurations between pairs enables a larger $\phi$.
                Each point corresponds to a pair of models $\theta_1,\theta_2$ that are fine-tuned independently
                from a shared initialization $\theta_0$ with different hyperparameter configurations.
                The angle $\phi$ between between solutions refers to the angle between
                $\theta_1 - \theta_0$ and $\theta_2 - \theta_0$ (i.e., the initialization is treated as the origin).
                Accuracy is averaged over ImageNet and the five distribution shifts described in Section~\ref{sec:setup}.}
                \label{fig:angles}
                            \vspace*{-0.2cm}
        \end{figure}
        
        \textbf{Error landscape visualizations.} To provide intuition, we visualize a two dimensional slice of the
        training loss and test error landscape when fine-tuning CLIP on ImageNet.
        In these experiments, we use the zero-shot initialization $\theta_0 \in \mathbb{R}^d$ and
        fine-tune twice, independently, to produce solutions $\theta_1$ and $\theta_2$.
        The points $\theta_0,\theta_1$ and $\theta_2$ define a plane in parameter space, and we evaluate the ImageNet train loss, ImageNet test error, and the test error on the five aforementioned
        distribution shifts on this plane. The results are illustrated
        in Figure~\ref{fig:error} where the zero-shot initialization ($\theta_0$) is shown as a star
        and a solution fine-tuned with learning rate $3\cdot 10^{-5}$ ($\theta_1$) is shown as a blue square.
        For $\theta_2$ we either use the same learning rate as $\theta_1$ (but vary the random seed) or learning rate $3 \cdot 10^{-6}$.
        For both the in-distribution and out-of-distribution test sets, the loss/error contours are basin-shaped, and none of the three points is optimal.\yair{edited the sentence to avoid the somewhat vague ``error basin''}
        
        These results suggest that (1) interpolating the weights of two fine-tuned solutions can improve accuracy
        compared to individual models and (2) more uncorrelated solutions---models that form an angle\footnote{In particular, the angle $\phi$ between $\theta_1 - \theta_0$ and
        $\theta_2 - \theta_0$, i.e., the angle between the arrows shown in Figure~\ref{fig:error}.} closer to 90 degrees---may lead to higher accuracy on the linear interpolation path.
        
        \begin{figure*}
            \centering
            \includegraphics[width=\textwidth]{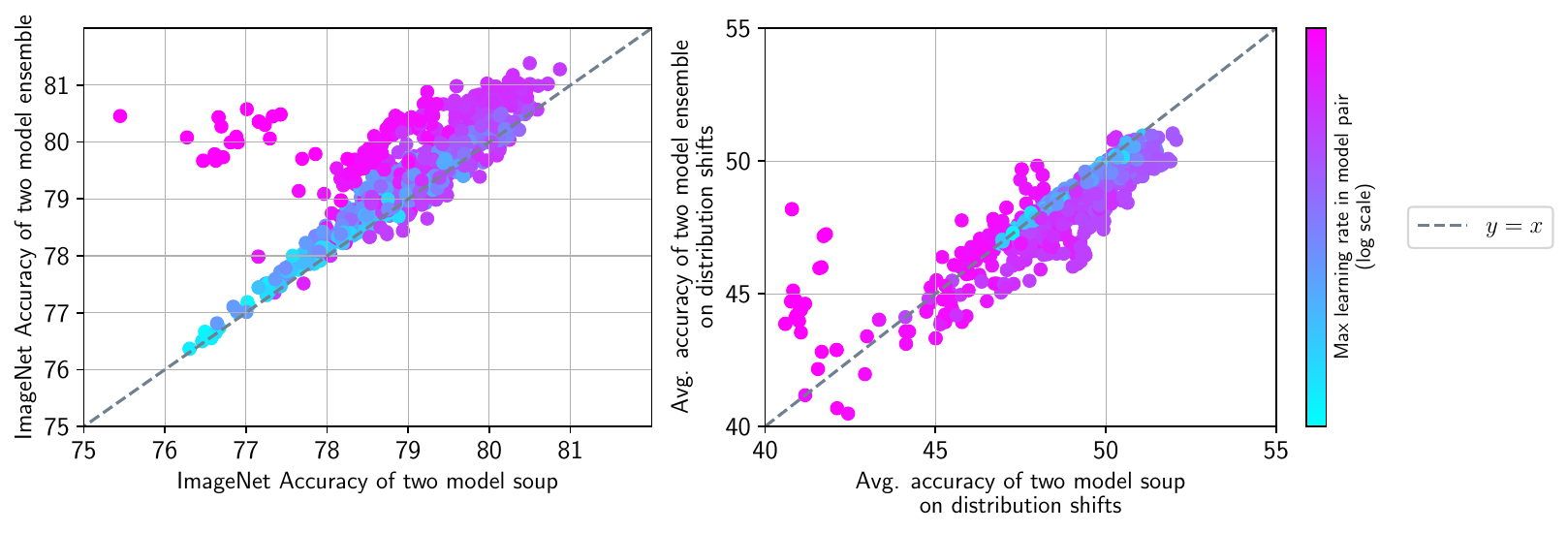}
            \vspace*{-0.7cm}
            \caption{Ensemble performance is correlated with model soup performance.
            Each point on the scatter plot is a model pair with different hyperparameters.
            The $x$-axis is the accuracy when the weights of the two models are averaged (i.e., the two model soup) while the $y$-axis is the accuracy of the two model ensemble.
            Ensembles often perform slightly better than soups on ImageNet (left) while the reverse is true on
            the distribution shifts (right).
            Each model pair consists of two random greed diamonds from Figure~\ref{fig:teaser}.}
            \label{fig:ose_wse_compare}
        \end{figure*}
        \begin{figure}
            \centering
            \includegraphics[width=0.9\columnwidth]{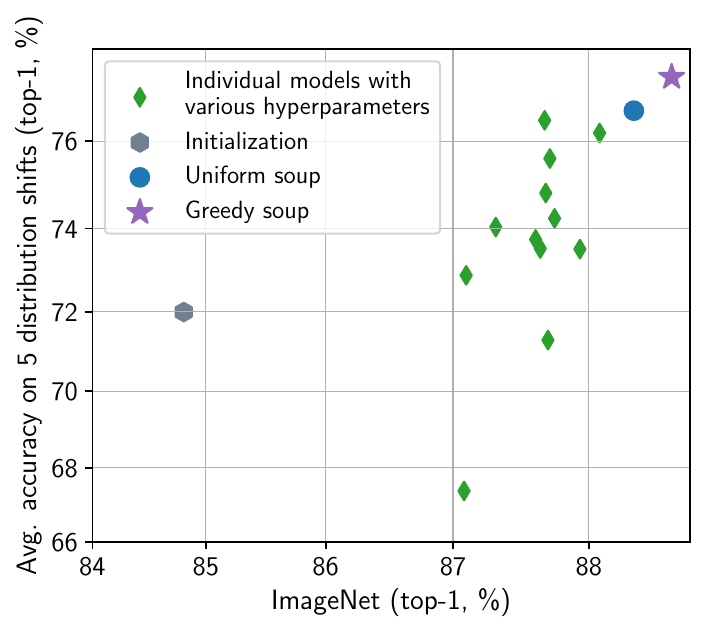}
            \caption{\emph{Model soups} improve accuracy when fine-tuning ALIGN.}
            \label{fig:align}
        \end{figure}

        To investigate the correlation between accuracy improvement and angle, 
        we consider a series of models trained with different seeds, learning rates, and data augmentation.
        For each pair $\theta_1,\theta_2$, we compare the accuracy of their average with the average of
        their accuracies, $\mathsf{Acc}\left(\frac{1}{2}\theta_1 +  \frac{1}{2} \theta_2\right) - \frac{1}{2} \left(  \mathsf{Acc}\left(\theta_1\right) +  \mathsf{Acc}\left(\theta_2\right) \right)$,
        which we refer to as the interpolation advantage. Figure~\ref{fig:angles} illustrates the results, in which we observe that the interpolation advantage is correlated with the angle $\phi$ and that varying the learning rate, seed, or data augmentation
        can produce solutions which are more orthogonal.
        Experimental details and discussion of high learning rates provided in Appendix~\ref{app:error}.

        \textbf{Ensemble comparison.} Figure~\ref{fig:ose_wse_compare} observes that ensemble performance is
        correlated with soup performance for moderate and small learning rates. We consider pairs of models selected at random from the individual solutions in Figure~\ref{fig:teaser}, and find that the maximum learning rate of the models in the pair is indicative of the ensemble accuracy, soup accuracy, and their relation:
        When learning rate is small, ensemble accuracy and soup accuracy are similar, but both are suboptimal.
        For moderate learning rate values, ensemble accuracy and soup accuracy are both high.
        For high learning rate values, ensemble performance exceeds soup performance, but ensembles/soups with
        moderate learning rates perform better.
        Overall, ensembles achieve higher accuracy on ImageNet while the reverse is true on the distribution shifts.
        
        \textbf{One dimensional hyperparameter grids.} 
        Finally, in Appendix~\ref{app:1d} we ask the question: for a one dimensional grid of hyperparameters $\{h_a,...,h_b\}$,
        how does averaging the models fine-tuned with
        hyperparameter configurations $h_a$ and $h_b$ corresponding to the endpoints compare with
        picking the best individual model fine-tuned with hyperparameter
        configuration $h \in \{h_a,...,h_b\}$?
        The hyperparameters we vary are optimizer,
        augmentation, and learning rate.
        For the majority of grid searches, the average of the endpoints outperforms
        the best individual model in the grid.
        
        \subsection{Model soups}\label{sec:mainres}
        
        With the gains of averaging two fine-tuned models in mind, we turn our attention to averaging \emph{many} models with different hyperparameters:  
        this section presents our main results, which show that averaging fine-tuned models can be used
        as an alternative to the conventional procedure of selecting the single
        model which performs best on the held-out validation set.
        We explore CLIP~\cite{radford2021learning} and ALIGN~\cite{jia2021scaling} fine-tuned on ImageNet~\cite{deng2009imagenet} (Section~\ref{sec:clipalign}), 
        ViT-G pre-trained on JFT-3B~\cite{zhai2021scaling} and fine-tuned on ImageNet (Section~\ref{sec:jft}), and transformer models fine-tuned on
        text classification tasks (Section~\ref{sec:nlp}).
        Appendix~\ref{app:moresets} additionally explores (1) CLIP ViT-L fine-tuned on WILDS~\cite{wilds2021} and CIFAR-10 and (2) an ImageNet-22k-pretrained ViT-B fine-tuned on ImageNet. %
        Moreover, Appendix~\ref{app:basic} shows that model soups improve accuracy when fine-tuning BASIC~\cite{pham2021scaling}.
    
    \subsubsection{Fine-tuning CLIP and ALIGN}\label{sec:clipalign}

        We begin our study of model soups by considering two-pretrained models, CLIP ViT-B/32 and ALIGN EfficientNet-L2, and performing a hyperparameter sweep for the fine-tuning each model on ImageNet. For CLIP we use a random hyperparameter search over learning rate, weight decay,
        training epochs, label smoothing, and data augmentation, obtaining 72 fine-tuned models (details in Appendix~\ref{app:clipdetails}).
        For ALIGN we use a grid search over learning rate, data augmentation, and mixup, obtaining 12 fine-tuned models (details in Appendix~\ref{app:aligndetails}).
        To form our greedy soups, we sort models in order of decreasing accuracy on the held-out validation set 
        before applying Recipe~\ref{alg:greedy}.
        For both CLIP and ALIGN, the greedy soup selects 5 models.
        Figure~\ref{fig:teaser} and \ref{fig:align} show the performance of the resulting models and their uniform and greedy soups for CLIP and ALIGN. The greedy soup improves on over the best model in the hyperparameter sweep by 0.7 and 0.5 percentage points, respectively.
        
        \begin{table}[t!]
            \caption{Ablation on multiple methods from Table~\ref{tab:methods} and their variants when 
            when fine-tuning CLIP ViT-B/32 with the random hyperparameter search described in Section~\ref{sec:clipalign}. 
            For ``Greedy soup (random order)'', we try three random model orders when running the greedy soup procedure
            (by default, models are sorted by decreasing held-out val accuracy).
            The ``Learned soup'' and its variants are descried in Appendix~\ref{app:learned-soup}.
            The \emph{best} in \emph{best individual model} refers to ImageNet accuracy.}
                        \vspace*{-0.2cm}
            \begin{center}\footnotesize
            \setlength{\arrayrulewidth}{.01em}
            \begin{tabular}{lll}
                \toprule
                {} &              ImageNet &            Dist. shifts\\
                \midrule
                Best individual model        &           80.38 &          47.83 \\
                Second best model &           79.89 &         43.87 \\
                \midrule
                Uniform soup                                 &           79.97 &          51.45 \\
                Greedy soup &           81.03 &          50.75 \\
                Greedy soup (random order)               &  $80.79$ $\scriptstyle(0.05)$ &  $51.30$ $\scriptstyle(0.16)$ \\
                Learned soup                             &           80.89 &          51.07 \\
                Learned soup (by layer)                  &           81.37 &          50.87 \\
                \midrule
                Ensemble                &           81.19 &          50.77 \\
                Greedy ensemble           &            81.90 &          49.44 \\
                \bottomrule
            \end{tabular}
                        \vspace*{-0.4cm}
            \end{center}
            \label{tab:results}
        \end{table}

        \begin{table*}
            \caption{Greedy soup improves over the best individual models obtained in a hyperparameter sweep for ViT-G/14 pre-trained on JFT-3B and fine-tuned on ImageNet, both in- and out-of-distribution. Accuracy numbers not significantly different from the best are bold-faced. Statistical comparisons are performed using an exact McNemar test or permutation test at $\alpha = 0.05$. Avg shift accuracy of the best model on each test set is the best average accuracy of any individual model.
            Analogous results when fine-tuning BASIC-L are available in Appendix~\ref{app:basic}.}
            \vspace*{-0.3cm}
            \label{tab:vit_g_finetuned_result}
            \centering
            \footnotesize
            \setlength{\tabcolsep}{0.4em}
            \begin{tabular}{ l | r r r | r r r r r | r}
                \toprule
                & \multicolumn{3}{c|}{ImageNet} & \multicolumn{5}{c|}{Distribution shifts} & \\\cmidrule(l{2pt}r{2pt}){2-4}\cmidrule(l{2pt}r{2pt}){5-9}
                \multicolumn{1}{c|}{Method} & Top-1 & ReaL & Multilabel  & \multicolumn{1}{c}{IN-V2} & \multicolumn{1}{c}{IN-R} & \multicolumn{1}{c}{IN-Sketch} & \multicolumn{1}{c}{ObjectNet} & \multicolumn{1}{c|}{IN-A} & \multicolumn{1}{c}{Avg shifts}\\
                \midrule
                ViT/G-14~\cite{zhai2021scaling} & 90.45 & 90.81 & -- & 83.33 & -- & -- & 70.53 & -- & -- \\
                CoAtNet-7~\cite{dai2021coatnet} & 90.88 & -- & -- & -- & -- & -- & -- & -- & -- \\
                \midrule
                \multicolumn{10}{l}{\textit{Our models/evaluations based on ViT-G/14:}}\\
                ViT/G-14~\cite{zhai2021scaling} (reevaluated) & 90.47 & 90.86 & 96.89 & 83.39 & 94.38 & 72.37 & 71.16 & 89.00 & 82.06 \\
               Best model on held out val set & 90.72 & 91.04 & 96.94 & 83.76 & 95.04 & 73.16 & 78.20 & 91.75 & 84.38 \\
               Best model on each test set (oracle) & 90.78 & \textbf{91.78} & \textbf{97.29} & \textbf{84.31} & 95.04 & 73.73 & \textbf{79.03} & 92.16 & 84.68 \\
                Greedy ensemble & \textbf{90.93} & 91.29 & \textbf{97.23} & \textbf{84.14} & 94.85 & 73.07 & 77.87 & 91.69 & 84.33\\
                Greedy soup & \textbf{90.94} & 91.20 & \textbf{97.17} & \textbf{84.22} & \textbf{95.46} & \textbf{74.23} & 78.52 & \textbf{92.67} & \textbf{85.02} \\
                
                \bottomrule
            \end{tabular}
            \end{table*}
                \begin{table*}
                    \caption{Performance of model soups on four text classification datasets from the GLUE benchmark \cite{wang2018glue}.}
                    \vspace*{-0.4cm}
                    \begin{center}\small
                    \setlength{\arrayrulewidth}{.01em}
                    \setlength{\tabcolsep}{12pt}
                    \begin{tabular}{llcccc}
                        \toprule
                        Model & Method & MRPC & RTE & CoLA & SST-2  \\\midrule
                        \multirow{2}{*}{BERT \cite{devlin2019bert}} & Best individual model & 88.3 & 61.0 & 59.1 & 92.5 \\
                        & Greedy soup & 88.3 {\textbf{\scriptsize\textcolor{gray}{(+0.0)}}} & 61.7 {\textbf{\scriptsize\textcolor{limegreen}{(+0.7)}}} & 59.1 {\textbf{\scriptsize\textcolor{gray}{(+0.0)}}} & 93.0 {\textbf{\scriptsize\textcolor{limegreen}{(+0.5)}}} \\\midrule
                        \multirow{2}{*}{T5 \cite{raffel2020t5}} & Best individual model & 91.8 & 78.3 & 58.8 & 94.6\\
                        &  Greedy soup & 92.4 {\textbf{\scriptsize\textcolor{limegreen}{(+0.6)}}} & 79.1 {\textbf{\scriptsize\textcolor{limegreen}{(+0.8)}}} & 60.2 {\textbf{\scriptsize\textcolor{limegreen}{(+0.4)}}} & 94.7 {\textbf{\scriptsize\textcolor{limegreen}{(+0.1)}}}\\
                        \bottomrule
                    \end{tabular}
                    \end{center}
                    \label{tab:nlp}
                    \vspace*{-0.4cm}
                \end{table*}

        Furthermore, we show that, for essentially any number of models, the greedy soup outperforms the best single model on both the ImageNet and the out-of-distribution test sets.
        We consider an additional setting where we prepare a sequence of soups by sequentially adding CLIP models from the hyperparameter sweep in random order.
        Appendix Figure~\ref{fig:active} shows the performance of the uniform and greedy soup, as well as the best single model so far and a logit ensemble, as a function of the number of models considered. The greedy soup is better than the uniform soup on ImageNet and comparable to it out-of-distribution. The logit ensemble is better than the greedy soup on ImageNet, but worse out-of-distribution.
        
        Table~\ref{tab:results} lists the performance of the CLIP soups and baselines described above, as well as additional soup variants described in Appendix~\ref{app:learned-soup}.
        
        To further establish the generality of the model soup, we replicate the CLIP hyperparameter sweep experiment on two image classification tasks from WILDS~\cite{wilds2021}, namely FMoW~\cite{christie2018functional} and iWildCam~\cite{beery2021iwildcam}. Appendix Figure~\ref{fig:fmow-iwc} shows results  qualitatively similar to our ImageNet experiment, and Appendix~\ref{app:clipdetails} describes experimental details. 
        
        We report several additional variants and baselines for the experiment described above. In Appendix~\ref{app:moreinits} we present results for different hyperparameter sweeps and fine-tuning initializations, when fine-tuning CLIP on ImageNet. 
        For instance, we try a $\emph{standard grid}$ search which is similar to the
        grid search described for ALIGN above, and an $\emph{extreme grid}$ search which includes
        solutions fine-tuned with extreme hyperparameters that result in badly performing models (details in Appendix~\ref{app:clipdetails}).
        Moreover, Appendix~\ref{app:baselines}
        compares model soups with additional
        baselines, including distillation from
        an ensemble as in~\citet{hinton2014dark},
        exponential moving averaging~\cite{szegedy2016rethinking}, stochastic weight averaging~\cite{izmailov2018averaging}, and sharpness aware
        minimization~\cite{foret2021sharpnessaware}.
        
        We highlight a few interesting takeaways from these experiments:
        (1) The greedy soup outperforms the best individual model---with no extra training and no extra compute
        during inference, we were able to produce a better model.
        (2) While the
        uniform soup can outperform the best individual model, we only observe this when all individual
        models achieve high accuracy (e.g., when fine-tuning ALIGN in Figure~\ref{fig:teaser}); 
        unlike the examples in Figure~\ref{fig:error}, there can be an error barrier between fine-tuned models.
        We mainly observe this when fine-tuning with high learning rates (this is illustrated in Appendix~\ref{app:error}, Figure~\ref{fig:errorbig}).
        However, these high learning rate models also have a lower accuracy,
        and are therefore excluded by the greedy soup.

    \subsubsection{Fine-tuning a ViT-G model pre-trained on JFT-3B}\label{sec:jft}
    \label{sec:vit-g}
    
    To test whether the gains obtained by model soups are additive with other techniques used to obtain state-of-the-art models, we applied our greedy soup technique to 58 ViT-G/14 models fine-tuned on ImageNet. We vary the learning rate, decay schedule, loss function, and minimum crop size in the data augmentation, and optionally apply RandAugment~\cite{cubuk2020randaugment}, mixup~\cite{zhang2017mixup}, or CutMix~\cite{yun2019cutmix}. We also train four models with sharpness-aware minimization (SAM)~\cite{foret2021sharpnessaware}. For further details of our hyperparameter sweep, see Appendix~\ref{app:vitgdetails}. For each model training run, we save exponential moving averages (EMA) of the weights~\cite{szegedy2016rethinking} computed with decay factors of 0.999 (low EMA) and 0.9999999 (high EMA). Whereas high EMA generally provides the best single-model accuracy, both greedy soup and greedy ensembling attain higher validation accuracy when applied to parameters with low EMA. We report the highest single model accuracy numbers obtained with either EMA decay value, but perform greedy soup and ensembling with models trained with EMA decay of 0.999. For each combination of training run and EMA decay rate, we evaluate accuracy on our held out validation set every 1000 steps. We use these accuracy values to pick the best checkpoint for ensembling, souping, and subsequent evaluation.
    
    In Table~\ref{tab:vit_g_finetuned_result}, we report results on the ImageNet validation set and the five distribution shift datasets studied above as well as two relabeled ImageNet validation sets, ReaL~\cite{beyer2020we} and multilabel~\cite{shankar2020evaluating}. Our greedy soup procedure selects 14 of the 58 models fine-tuned as part of our hyperparameter sweep, and this soup performs statistically significantly better than the best individually fine-tuned model selected based on our held out validation set on all datasets except for ObjectNet. Even when we give an unfair advantage to individually fine-tuned models by selecting them based on their performance on each test set (denoted ``oracle'' in Table~\ref{tab:vit_g_finetuned_result}), the greedy soup, which was selected using only in-distribution data, remains superior on most datasets. Only on ReaL and ObjectNet does there exist an individual model that performs statistically significantly better than the soup, and the best model differs between those two datasets. Greedy ensembling performs similarly to the greedy soup in terms of ImageNet top-1 and multilabel accuracy, and slightly better on ReaL, but significantly worse on all distribution shift datasets except for ImageNet-V2. Thus, greedy soup can provide additional gains on top of standard hyperparameter tuning even in the extremely high accuracy regime.%
        
        \subsubsection{Fine-tuning on text classification tasks}\label{sec:nlp}
    
        To test whether the gains obtained by model soups extend to domains beyond image classification, we conduct preliminary experiments with natural language processing (NLP). 
        While more investigation is warranted
        to establish the applicability of model soups for NLP, we believe our experiments are a promising initial step.
        In particular, we fine-tune BERT \cite{devlin2019bert} and T5 \cite{raffel2020t5} models on four text classification tasks from the GLUE benchmark \cite{wang2018glue}: MRPC \cite{dolan2005automatically}, RTE \cite{dagan2005pascal, bar2006second, giampiccolo2007third, bentivogli2009fifth}, CoLA \cite{warstadt2018neural} and SST-2 \cite{socher2013recursive}, as in \cite{dodge2020fine}. We use the standard metric for each dataset: average of accuracy and $F_1$ score for MRPC, accuracy for RTE, Matthews correlation for CoLA \cite{matthews1975comparison} and accuracy for SST-2. Details are provided in Appendix \ref{app:nlp_datasets}.
        
        We fine-tune 32 models for each dataset with a random hyper-parameter search over learning rate, batch size, number of epochs and random seed. Table~\ref{tab:nlp} reports the corresponding metric on the validation set for BERT-base uncased \cite{devlin-etal-2019-bert} and T5-base \cite{raffel2020t5}. Additional experimental details and results for more models are provided in Appendix \ref{app:nlp_ft}. While the improvements are not as pronounced as in image classification, the greedy soup can improve performance over the best individual model in many cases.

        \section{Analytically comparing soups to ensembles}\label{sec:theory}
          
        \newcommand{\grad}{\nabla}%
        \newcommand{\hess}{\nabla^{2}}%
        \newcommand{\R}{\mathbb{R}}%
        \newcommand{\E}{\mathbb{E}}%
        \newcommand{\err}{\mathrm{err}}
        \newcommand{\errens}{\mathrm{err}^{\mathrm{ens}}}
        \newcommand{\fose}{f_\alpha^{\mathrm{ens}}}%
        \newcommand{\fwse}{f_\alpha^{\mathrm{soup}}}%
        \newcommand{\defeq}{\coloneqq}%
        \newcommand{\indic}[1]{1{\{#1\}}}%
        \newcommand{\exloss}{\mathcal{L}^{\mathrm{soup}}}
        \newcommand{\exlossens}{\mathcal{L}^{\mathrm{ens}}}
        \renewcommand{\d}{\mathrm{d}}
        \newcommand{\softmax}{p_\mathrm{sftmx}}
        
        The goal of this section is to obtain complementary analytical insight into the effectiveness of model soups. For simplicity, we consider a soup consisting of only two models with parameters $\theta_0$ and $\theta_1$. For weighting parameter $\alpha\in[0,1]$ we let
        $
        \theta_\alpha = (1-\alpha)\theta_0 + \alpha \theta_1
        $	
        denote the weight-averaged soup. We would like to understand when the soup error, $\err_\alpha \defeq \E_{x,y} \indic{\argmax_i f_i(x;\theta_\alpha)\ne y}$, would be lower that the best of both endpoints, $\min\{\err_0, \err_1\}$.
        
        Note that just convexity of $\err_\alpha$ in $\alpha$ does not by itself imply superiority of the soup to both endpoints, as the minimum of $\err_\alpha$ over $\alpha$ may be obtained at the endpoints even when $\err_\alpha$ is convex. To get further leverage on the problem, we compare the soup to the \emph{logit-level ensemble}
        $
            \fose(x) = (1-\alpha)f(x;\theta_0) + \alpha f(x;\theta_1).
        $
        The rich literature on ensembles (see Sec.~\ref{sec:related}) tells us that the expected error of the ensemble, $\errens_\alpha$, is often strictly below $\min\{\err_0, \err_1\}$ for neural networks. Therefore, whenever $\err_\alpha \approx \errens_\alpha$ we expect the soup to outperform both endpoint models.
        
        To analytically compare the soup and the ensemble, we replace the 0-1 loss with a differentiable surrogate. Specifically, we consider the cross-entropy loss $\ell(f,y)=\log\left(\sum_{y'} e^{f_{y'} - f_y}\right)$. We let $\exloss_\alpha = \E_{x,y} \ell(\beta f(x;\theta_\alpha),y)$ denote the $\beta$-calibrated expected loss of the soup, and similarly define $\exlossens_\alpha = \E_{x,y} \ell(\beta \fose(x),y)$ for the ensemble. We derive the following approximation for the loss difference:
        \begin{flalign}
            & \exloss_\alpha - \exlossens_\alpha
            \approx 
            \frac{\alpha(1-\alpha)}{2} \bigg( 
            -\frac{\d^2}{\d \alpha^2} \exloss_\alpha \nonumber\\& ~~~~~~
            + \beta^2 \E_{x} \mathrm{Var}_{Y\sim \softmax(\beta f(x;\theta_\alpha))} \left[ \Delta f_Y(x) \right]
            \bigg),
            \label{eq:approx}
        \end{flalign}
        where $[\softmax(f)]_i = e^{f_i} / \sum_{j} e^{f_j}$ is the standard ``softmax'' distribution and $\Delta f(x) = f(x;\theta_1)-f(x;\theta_0)$ is the difference between the endpoint logits. We obtain our approximation in the regime where the logits are not too far from linear; see Appendix~\ref{app:theory-deriv} for a detailed derivation.
        
        The first term in approximation~\eqref{eq:approx} is negatively proportional to the second derivative of the loss along the trajectory: when the approximation holds, convexity of the loss indeed favors the soup. However, the second term in the approximation does not follow from the ``convex basin'' intuition. This term always favors the ensemble, but is small in one of two cases: (a) the somewhat trivial case when the endpoint models are similar (so that $\Delta f$ is small) and (b) when the soup produces confident predictions, implying that $\softmax(\beta f(x;\theta_\alpha))$ is close to a point mass and consequently the variance term is small.
        
        To test our approximation, we evaluate it over of set of fine-tuned models 
        with different learning rates, augmentation strategies, random seeds and $\alpha$ values. We set $\beta$ to calibrate the soup model, and find that it improves the ability of our approximation to predict the soup/ensemble error difference; see Appendix~\ref{app:theory-eval} for detailed description of our  setup.
        
        Figure~\ref{fig:theory-eval} summarizes the results of our empirical evaluations. When excluding the high learning rate of $10^{-4}$ (center and right panels),\footnote{Fine-tuned models with learning rate $10^{-4}$ are far in weight space from the initial model and are often rejected when forming greedy soups. %
        Therefore, we do not expect our approximation to be tight for these learning rates.} we see that the approximation is strongly correlated with both the true difference in loss as well as the difference in error, and the approximation and true loss difference generally agree in sign.
        Additional details are provided in Appendix~\ref{app:theory}.

        \section{Scope and limitations}\label{sec:lim}
        
        While this work has so far demonstrated that
        averaging many fine-tuned models is a useful
        technique for improving accuracy,
        this section explores two limitations of the approach.
        The first is the applicability of model soups,
        and the second is the failure of model soups
        to substantially improve calibration.
        
        \textbf{Applicability.} 
        So far our experiments have mainly explored
        models pre-trained on large, heterogeneous datasets.
        In Appendix~\ref{app:moresets} we also explore
        model soups for an ImageNet-22k pre-trained
        model.
        While the greedy soup still provides
        improvements on ImageNet, these improvements are less substantial compared to those observed when fine-tuning CLIP and ALIGN.

        \textbf{Calibration.}
        While ensembles improve model calibration~\cite{guo2017calibration, roelofs2020mitigating},
        model soups do not have the same effect.
        As hyperparameters can also have an effect on calibration,
        we consider the ensemble and soup of 20 models which are identical
        other than random seed.
        Results are illustrated in Figure~\ref{fig:cal} using the calibration
        metrics of \citet{roelofs2020mitigating}.
        
        \section{Related work}\label{sec:related}
        \vspace{-0.25em}
        
        \paragraph{Averaging model weights.}
        Averaging the weights of models is a popular approach
        in convex optimization and deep learning.
        Most applications study models
        along the same optimization trajectory, e.g.
        \cite{ruppert1988efficient,polyak1990new,szegedy2016rethinking,izmailov2018averaging,zhang2019lookahead, kaddour2022questions, junczys2016amu}.
        By contrast, \citet{nagarajan19uniform, frankle2020linear,neyshabur2020being, von2020neural} and \citet{matena2021merging} weight-average
        models which share an initialization but are optimized independently.
        \citet{nagarajan19uniform} observed that models trained on MNIST~\cite{lecun1998mnist} from the same random initialization are connected in weight space by a linear path of high accuracy.
        \citet{frankle2020linear} find that, when training a pair of models from scratch on harder datasets such as ImageNet with the 
        same hyperparameter configuration and initialization but different data order, interpolating weights
        achieves no better than random accuracy. However, \citet{frankle2020linear} showed that when the two models
        share a portion of their optimization trajectory, accuracy does not drop when
        they are averaged.
        Analogously, \citet{neyshabur2020being} demonstrate that when two
        models are fine-tuned with the same pre-trained initialization, the interpolated model attains
        at least the accuracy of the endpoints.
        Unlike \citet{nagarajan19uniform, frankle2020linear,neyshabur2020being} we consider averaging many models
        with varied hyperparameter configurations.
        
        In the late phases of training, \citet{von2020neural} make copies of a subset of the neural network parameters (e.g, the batch norm weights, the classification layer, etc.). These parameters are then optimized independently and subsequently averaged. In contrast to \citet{von2020neural}, a) we average across independent runs with hyperparemter diversity, b) we modify all weights in the network, and c) we consider the transfer setting.
        \citet{matena2021merging} merge models with the same pre-trained initialization that are
        fine-tuned on different text classification tasks.
        They also propose Fisher information as an alternative technique for model merging.
        We experiment with averaging models which are trained on different datasets in Appendix~\ref{sec:zsperf}, however, in contrast to \citet{matena2021merging} we do not use data from the target distribution.
        \citet{wortsman2021robust} average zero-shot and fine-tuned models, finding improvements in- and out-of-distribution.
        In contrast to \citet{wortsman2021robust}, we average models across many independent runs which provides more substantial improvements.
        
        Stochastic Weight Averaging (SWA)~\cite{izmailov2018averaging}, which averages weights along a single optimization trajectory, is also motivated by the relation between ensembling model outputs and averaging model weights. In contrast, the averaging we propose is across independent runs. Moreover, while their analysis relates the averaged network outputs (i.e., the logit ensemble) to the output of the a network with the averaged weights, our analysis (Section~\ref{sec:theory}) goes a step further and relates the classification losses associated with these two vectors.
        
        \paragraph{Pre-training and fine-tuning.}
        In computer vision and natural language processing, the best performing models are often pre-trained on a
        large dataset before being fine-tuned on
        data from the target task
        \cite{donahue2014decaf,yosinski2014transferable, sharif2014cnn, girshick2014rich, mahajan2018exploring,kornblith2019better, yalniz2019billion, kolesnikov2020big,bommasani2021opportunities}.
        This paradigm is also referred to as transfer learning.
        Recently, image-text pre-training has become 
        increasingly popular in computer vision as a pre-training task
        \cite{radford2021learning, jia2021scaling, mu2021slip, pham2021scaling, yu2022coca}.
        Recent work has explored alternative strategies for adapting these models to specific target tasks \cite{coop,gao2021clip,zhang2021tip}, for instance via a lightweight residual feature adapter.
        In contrast, our work explores standard end-to-end fine-tuned models. Other work has attempted to improve transfer learning by regularizing models toward their initialization~\cite{xuhong2018explicit}, choosing layers to tune on a per-example basis~\cite{guo2019spottune}, reinitializing layers over the course of training~\cite{li2020rifle}, or using multiple pretrained models with data-dependent gating~\cite{shu2021zoo}.

        \paragraph{Ensembles.} Combining the outputs of many models is a foundational
        technique for improving the accuracy and robustness of machine learning models 
        \cite{dietterich2000ensemble, bauer1999empirical, breiman1996bagging,friedman2001elements, deepensembles, FREUND1997119}.
        \citet{ovadia2019can} show that ensembles exhibit high accuracy under distribution shift.
        \citet{mustafa2020deep} propose a method for identifying subsets of pre-trained models for fine-tuning and later ensembling them, finding strong in-distribution accuracy and robustness to distribution shift.
        \citet{gontijo2021no} conduct a large-scale study of ensembles, finding that higher divergence in training methodology leads to uncorrelated errors and better ensemble accuracy.
        Finally, previous work has explored building ensembles of models produced by hyperparameter searches \cite{snoek2015scalable,mendoza2016towards,saikia2020optimized}, including greedy selection strategies \cite{caruana2004ensemble,caruana2006getting,levesque2016bayesian,wenzel2020hyperparameter}.
        Importantly, ensembles require a separate inference pass through each model, which increases computational costs. When the number of models is large, this can be prohibitively expensive. 
        Unlike ensembles, model soups require no extra compute at inference time.
        
        \vspace{-0.5em}
        \section{Conclusion}
        \vspace{-0.25em}
        Our results challenge the conventional procedure of selecting the best model on the held-out validation set when fine-tuning.
        With no extra compute during inference, we are often able to produce a better model
        by
        averaging the weights of multiple fine-tuned solutions.
    
    \subsection*{Acknowledgements}
    {
    We thank
    Ting Chen, Jesse Dodge, Ben Eysenbach, David Fleet, Pieter-Jan Kindermans, Mohammad Norouzi, Sarah Pratt and Vivek Ramanujan for helpful discussions and draft feedback, Lucas Beyer and Xiaohua Zhai for assistance with ViT-G/14 fine-tuning, 
    and Hyak at UW for computing support. 
    YC was supported in part by the Israeli Science Foundation (ISF) grant no.\ 2486/21, the Len Blavatnik and the Blavatnik Family foundation, and The Yandex Initiative for Machine Learning.
    This work is in part supported by the NSF AI Institute for Foundations of Machine Learning (IFML), Open Philanthropy, NSF IIS 1652052, IIS 17303166, DARPA N66001-19-2-4031, DARPA W911NF-15-1-0543 and gifts from Allen Institute for AI.
    }
    {\small
    \bibliographystyle{plainnat}
    \bibliography{main}
    }

    \newpage
    \onecolumn
    \appendix
    \counterwithin{figure}{section}
    \counterwithin{table}{section}

    \section{Overview}
    The appendix is organizes via the following contributions:
    \begin{itemize}
        \item Appendix~\ref{app:morefigs} (Additional figures) supplements the main text with additional figures.
        \item Appendix~\ref{app:basic} (BASIC) presents additional experiments exploring model soups for BASIC~\cite{pham2021scaling}.
        \item Appendix~\ref{sec:robustft} (Robust fine-tuning) compares model soups with WiSE-FT~\cite{wortsman2021robust}, a technique for fine-tuning while preserving robustness.
        \item Appendix~\ref{sec:zsperf} (Cross-dataset soups) explores soups for models which are trained on different datasets to improve zero-shot transfer.
        \item Appendix~\ref{app:1d} (Analysis of 1D hyperparameter grids) compares the performance of averaging endpoints with intermediate solutions for hyperparemters on a one dimensional grid.
        \item Appendix~\ref{app:moresets} (Additional fine-tuning and pre-training datasets) explores model soups for additional datasets.
        \item Appendix~\ref{app:moreinits} (Additional grid searches and initializations) supplements the results in the main text with other hyperparameter sweeps and model initializations (i.e., zero-shot instead of LP initialization).
        \item Appendix~\ref{app:learned-soup} (Learned soup) describes the more advanced souping procedure where we learn the soup mixing coefficients with gradient based optimization on the held-out validation set.
        \item Appendix~\ref{app:details} (Experimental details) provides additional details for the experiments.
        \item Appendix~\ref{app:theory} (Analytical comparison details) supplements Section~\ref{sec:theory} in analytically comparing soups and ensembles.
        \item Appendix~\ref{app:baselines} (Additional baselines) compares soups with additional baselines including stochastic weight averaging~\cite{izmailov2018averaging} and sharpenss aware minimization~\cite{foret2021sharpnessaware}.
    \end{itemize}
    
    \section{Additional figures}\label{app:morefigs}
    \begin{figure*}[h]
        \centering
        \includegraphics[width=\textwidth]{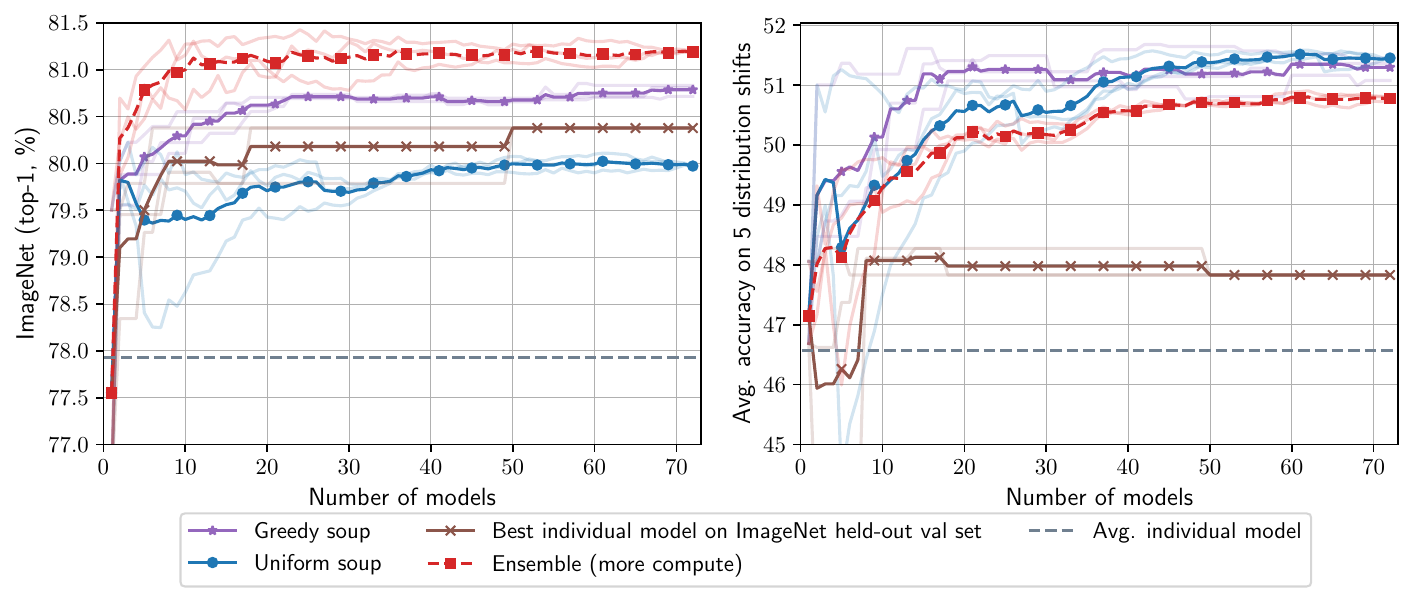}
        \caption{
        For essentially any number of models, the greedy soup outperforms the best single model on both ImageNet and the out-of-distribution test sets.
        On the $x$-axis we show the number of models considered in a random search over hyperparameters
        while the $y$-axis displays the accuracy of various methods for model selection which are summarized in Table~\ref{tab:methods}.
        All methods require the same amount of training and compute cost during inference with the exception of the ensembles,
        which require a separate pass through each model.
        Results are for fine-tuning CLIP ViT-B/32,
        averaged over three random orders (shown with faded lines).
        }
        \label{fig:active}
    \end{figure*}
    
    \begin{figure*}[h]
        \centering
        \includegraphics[width=\textwidth]{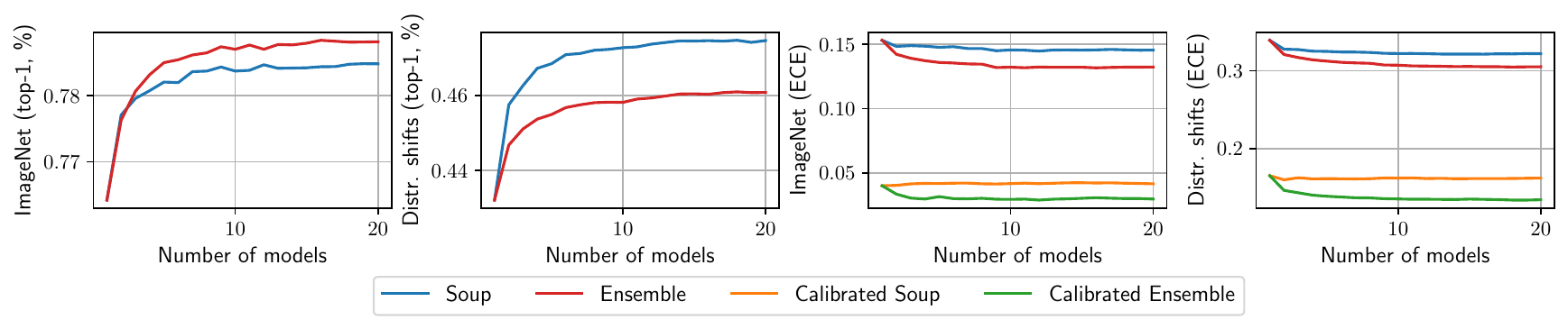}
        \vspace*{-0.8cm}
        \caption{Like model ensembling, model soups improve accuracy, but unlike model ensembling, model soups do not improve calibration. Expected calibration error (ECE) is computed using equal-mass binning.
        The soup in this figure is the uniform soup, and all models in this experiment are fine-tuned CLIP ViT-B/32 models
        with the same hyperparameters but different random seeds.
        The calibrated soup and calibrated ensemble refer to a soup and ensemble composed 
        of models which are calibrated through temperature scaling~\cite{guo2017calibration}.
        Calibrating models before ensembling or souping had no effect on accuracy and so these curves are omitted from the plots on the left.}
        \label{fig:cal}
    \end{figure*}
    \FloatBarrier
    \section{BASIC}\label{app:basic}
    After our initial submission we tested model soups when fine-tuning BASIC-L~\cite{pham2021scaling}. Due to memory constraints, we fine-tune with a batch size of 64 instead of 512. We initialize with the zero-shot classification head and train for 8 epochs using the Adafactor optimizer~\cite{shazeer2018adafactor} at a resolution of $500 \times 500$. We sweep over a grid of learning rates ($1 \cdot 10^{-5}$ or $2 \cdot 10^{-5}$) and 10 data augmentation settings, resulting in 20 different models. We use random crops and flips with a minimum crop size of 90\% of the image together with mixup~\cite{zhang2017mixup} or CutMix~\cite{yun2019cutmix} with $\alpha \in \{0.2, 0.4\}$, AutoAugment with $\verb|(num_layers, magnitude)| \in \{(2,10), (2, 15), (2, 20), (2, 25), (3, 10)\}$. We additionally train models with random crops and flips with minimum crop sizes of 5\% and 90\% without additional augmentation.
    
    As in our ViT-G/14 experiments (Section~\ref{sec:vit-g}), we save exponential moving averages with low and high EMA decay factors, and find that low EMA weights provide better performance for greedy souping and greedy ensembling whereas high EMA weights provide better single-model performance. We adjust the EMA factors for the difference in batch size and thus use a decay factor of $0.999^{1/8}$ for our low EMA configuration and $0.9999999^{1/8}$ for our high EMA configuration. During each training run, for each set of EMA weights, we evaluate accuracy on our held out validation set every 5000 steps and use the best checkpoint for ensembling, souping, and subsequent evaluation. We resize the full image to $500 \times 500$ for evaluation.
    
    Results are shown in Table~\ref{tab:basic_finetuned_result}. 
    The greedy soup consistently outperforms the individual model with highest accuracy on the held-out validation set. The best BASIC-L model on each individual test set sometimes outperforms the greedy soup, but selecting the model on the test set will generally overestimate its true accuracy.
    
            \begin{table*}[h]
            \caption{Greedy soup improves over the best individual model on the held-out validation set when fine-tuning BASIC-L~\cite{pham2021scaling}. Among the best model on the held out val set, the greedy ensemble, and the greedy soup, numbers not significantly different from the best are bold-faced. Statistical comparisons are performed using an exact McNemar test or permutation test at $\alpha = 0.05$. Avg shift accuracy of the best model on each test set is the best average accuracy of any individual model. For CoCa~\cite{yu2022coca}, a model which was introduced after our initial submission, evaluations were only available to one decimal place.}
            \label{tab:basic_finetuned_result}
            \centering
            \footnotesize
            \setlength{\tabcolsep}{0.4em}
            \begin{tabular}{ l | r r r | r r r r r | r}
                \toprule
                & \multicolumn{3}{c|}{ImageNet} & \multicolumn{5}{c|}{Distribution shifts} & \\\cmidrule(l{2pt}r{2pt}){2-4}\cmidrule(l{2pt}r{2pt}){5-9}
                \multicolumn{1}{c|}{Method} & Top-1 & ReaL & Multilabel  & \multicolumn{1}{c}{IN-V2} & \multicolumn{1}{c}{IN-R} & \multicolumn{1}{c}{IN-Sketch} & \multicolumn{1}{c}{ObjectNet} & \multicolumn{1}{c|}{IN-A} & \multicolumn{1}{c}{Avg shifts}\\
                \midrule
                ViT/G-14~\cite{zhai2021scaling} & 90.45 & 90.81 & -- & 83.33 & -- & -- & 70.53 & -- & -- \\
                CoAtNet-7~\cite{dai2021coatnet} & 90.88 & -- & -- & -- & -- & -- & -- & -- & -- \\
                BASIC-L (zero-shot)~\cite{pham2021scaling} & 85.7\textcolor{white}{0}  & -- & -- & 80.6\textcolor{white}{0} & 95.7\textcolor{white}{0} & 76.1\textcolor{white}{0} & 82.3\textcolor{white}{0} & 85.6\textcolor{white}{0} & 84.06 \\
                CoCa (zero-shot)~\cite{yu2022coca} & 86.3\textcolor{white}{0} & -- & -- & 80.7\textcolor{white}{0} & 96.5\textcolor{white}{0} & 77.6\textcolor{white}{0} & 82.7\textcolor{white}{0} & 90.2\textcolor{white}{0} & 85.54 \\
                CoCa (fine-tuned)~\cite{yu2022coca} & 91.0\textcolor{white}{0} & -- & -- & -- & -- & -- & -- & -- & -- \\
                ViT-G/14 greedy soup (Table~\ref{tab:vit_g_finetuned_result}) & 90.94 & 91.20 & 97.17 & 84.22 & 95.46 & 74.23 & 78.52 & 92.67 & 85.02 \\
                \midrule
                \multicolumn{10}{l}{\textcolor{gray}{\textit{Our models/evaluations with fine-tuned BASIC-L:}}}\\
                Best model on held out val set & 90.83  & 90.84 & 98.16 & \textbf{84.42} & 95.50 & 76.98 & 78.09 & 93.13 & 85.63 \\
                Greedy ensemble & \textbf{91.02}  & \textbf{91.11} & \textbf{98.46} & \textbf{84.65} & 95.79 & 76.63 & \textbf{79.91} & \textbf{94.05} & 86.20 \\
                Greedy soup & \textbf{90.98}  & 91.03 & 98.37 & \textbf{84.63} & \textbf{96.10} & \textbf{77.18} & \textbf{79.94} & \textbf{94.17} & \textbf{86.40} \\
                \midrule
                Best model on each test set (oracle) & 90.87  & 91.24 & 98.41 & 84.84 & 95.89 & 77.30 & 80.94 & 94.47 & 86.54 \\
                \bottomrule
            \end{tabular}
            \end{table*}

    \section{Robust fine-tuning}
    \label{sec:robustft}
        \begin{figure*}[t]
        \centering  
        \includegraphics[width=\textwidth]{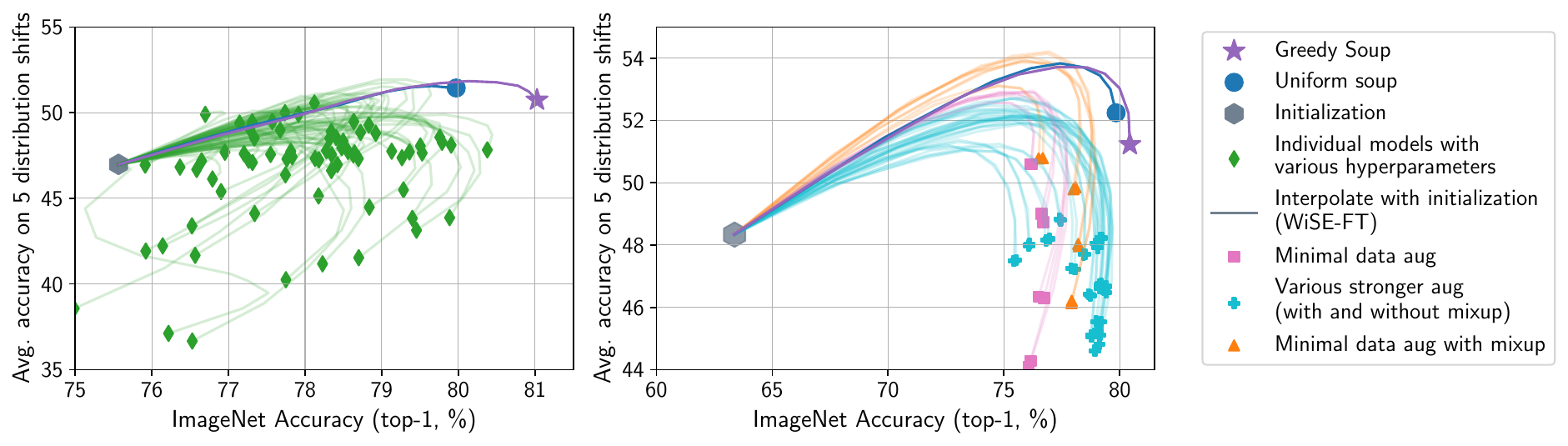}
        \vspace*{-0.7cm}
        \caption{
        Model soups compared to baselines for robust fine-tuning.
        WiSE-FT \cite{wortsman2021robust} improves the robustness of model $\theta_1$
        fine-tuned from initialization $\theta_0$ by interpolating between $\theta_1$ and $\theta_0$.
        Above we display the accuracy of models along these interpolation curves both for regular fine-tuned
        models and model soups (\textbf{left:} random hyperparameter search using the LP initialization, 
        \textbf{right:} grid search using the zero-shot initialization).
        The model soups lie beyond the WiSE-FT curves generated by any individual model,
        and accuracy can be improved on the distribution shifts by applying WiSE-FT to the
        model soups.
            \vspace*{-0.3cm}
        }
        \label{fig:scatter}
    \end{figure*}

    \citet{wortsman2021robust} introduce WiSE-FT, a method for improving the
    robustness of a model $\theta_1$ which is fine-tuned from initialization $\theta_0$ by linearly interpolating $\theta_1$ and $\theta_0$. 
    An intriguing observation was that, once the data augmentation is fixed, 
    interpolating between $\theta_1$ and $\theta_0$
    often traces a similar curve regardless of hyperparameters.\footnote{
    This is visible in Figure~\ref{fig:scatter} (right) where different data augmentations
    are shown with different colors. On the other hand, in Figure~\ref{fig:scatter} (left) there are many different methods
    of data augmentation as we conduct a random hyperparameter search.}
    In other words, a reasonable hypothesis was that this
    curve is Pareto optimal---no hyperparameter configuration would surpass it.
    In Figure~\ref{fig:scatter}, we trace the curves when interpolating between
    $\theta_1$ and $\theta_0$ for a random hyperparameter search (left) and the
    standard grid search described in Appendix~\ref{app:clipdetails} (right) when fine-tuning CLIP ViT-B/32.
    We find that the uniform soup and greedy soup lie
    beyond these interpolation curves. Moreover, we find 
    interpolating between these
    soups and the initialization also provides additional accuracy improvements on the distribution shifts.

    \section{Cross-dataset soups}\label{sec:zsperf}
    So far, our experiments have studied soups of models fine-tuned on the same dataset with different hyperparameters.
    In this section, we prepare soups containing models fine-tuned on different datasets.
    We evaluate the resulting soups on a held-out dataset, from which no labeled training data is used (i.e., zero-shot evaluation).
    
\begin{figure}[t]
        \centering
        \includegraphics[width=\columnwidth]{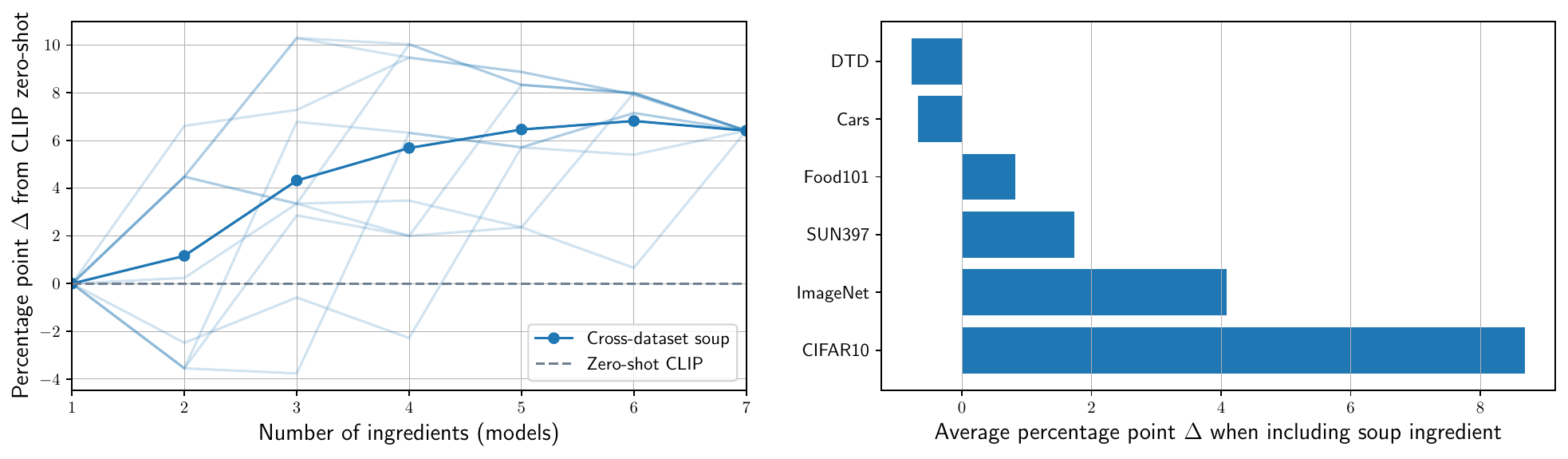}
        \caption{
        Model soups can improve zero-shot performance on new downstream tasks. 
        \textbf{(left)} Starting with zero-shot CLIP we create a soup by adding models fine-tuned on ImageNet, CIFAR-10, Food101, SUN397, DTD, and Cars,
        and evaluate on CIFAR-100.
        Different orders for adding models are shown with faded lines.
        \textbf{(right)} The average change in CIFAR-100 accuracy when a model fine-tuned on the dataset listed in the $y$-axis is added to the model soup.
        }
        \label{fig:samir}
    \end{figure}

    Concretely, we consider soups based on the CLIP zero-shot initialization along with six models fine-tuned independently on
    CIFAR-10 \cite{krizhevsky2009learning}, 
    Describable Textures \cite{dtd}, 
    Food-101 \cite{food101}, 
    SUN397 \cite{sun397}, 
    Stanford Cars \cite{cars} and ImageNet \cite{deng2009imagenet}.
    We evaluate on CIFAR-100 \cite{krizhevsky2009learning}, which does not share classes with CIFAR-10.
    Since each task has a different set of classes, the last layers cannot be part of the soup.
    Hence, during fine-tuning, we freeze the linear head produced by CLIP’s text tower so that task-specific learning is captured only in the backbone weights.
    At test time, we use the ``backbone soup'' with a zero-shot head constructed from CLIP's text tower and the CIFAR-100 class names with the prompt-ensembling used for ImageNet by~\citet{radford2021learning}.
    Figure~\ref{fig:samir} (left) shows that a model soup containing models trained on each of these
    datasets and the zero-shot model improves zero-shot performance on CIFAR-100 by
    6.4 percentage points over the CLIP baseline.
    Moreover, Figure~\ref{fig:samir} (right) shows that
    the choice of which fine-tuned models to include can have a substantial impact on the accuracy of the resulting soup.
    See Appendix~\ref{app:crossdataset} for additional details.

    \section{Analysis of 1D hyperparameter grids}\label{app:1d}
    
    This section asks: for a one dimensional grid of hyperparameters $\{h_a,...,h_b\}$,
    how does averaging the models fine-tuned with
    hyperparameter configurations $h_a$ and $h_b$ corresponding to the endpoints compare with
    picking the best individual model fine-tuned with hyperparameter
    configuration $h \in \{h_a,...,h_b\}$?
    \begin{figure*}[h!]
        \centering
        \includegraphics[width=\textwidth]{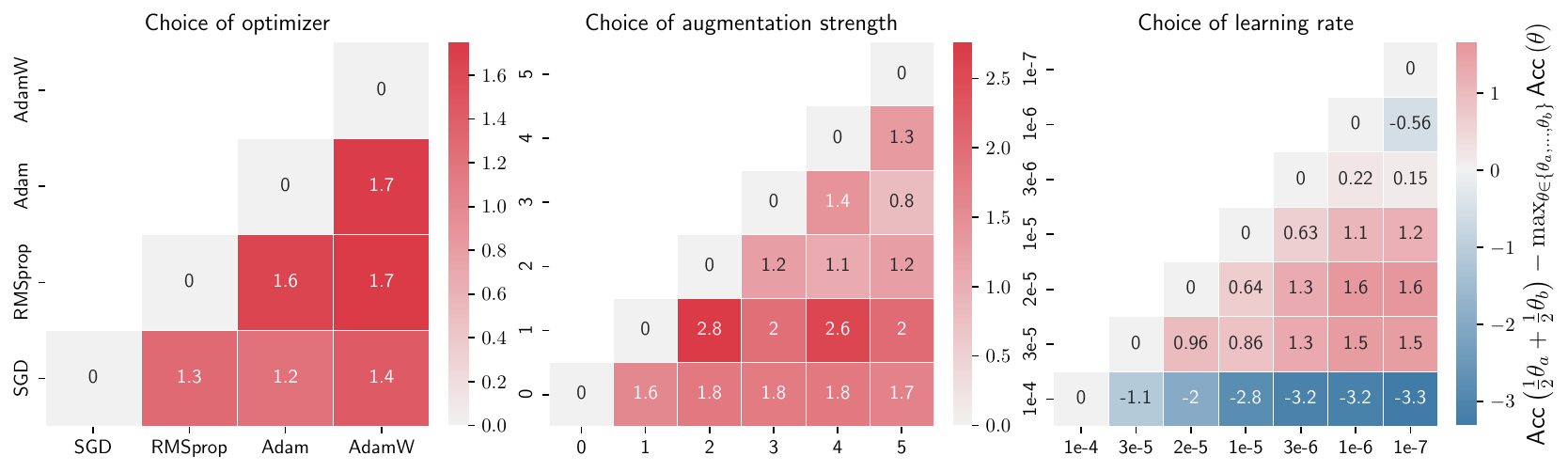}
        \caption{Analysis of 1D hyperparameter grids, where the average of models at the endpoints often outperforms
    the best individual model in the grid. In particular, colors and numbers indicate the percentage point improvement obtained by averaging the models on the $x$ and $y$ axis versus taking the best individual model in the range between them.
    Results are shown for the CLIP ViT-B/32 model fine-tuned on ImagNet.}
        \label{fig:grid}
    \end{figure*}
    
    The results are illustrated in Figure~\ref{fig:grid}, where each square represents a grid $\{h_a,...,h_b\}$.
    The average of the endpoints often outperforms
    the best individual model in the grid.
    A notable exception is when the learning rate $10^{-4}$ is the left endpoint of the grid.
    As this experiment uses AdamW, this learning rate is too high for fine-tuning and,
    unlike the examples in Figure~\ref{fig:error}, there is a high error barrier between
    the two fine-tuned solutions (see Figure~\ref{fig:errorbig}, lower right for example).
    
    When varying optimizer we use minimal data augmentation and LR $3\cdot 10^{-5}$
    for RMSProp~\cite{rmsprop}, Adam~\cite{kingma2014adam}, and AdamW~\cite{loshchilov2018decoupled}. SGD requires a larger learning rate, and so we use $0.1$.
    When varying augmentation strength, we use minimal data augmentation and LR $3\cdot 10^{-5}$.

    \section{Additional fine-tuning and pre-training datasets}\label{app:moresets}
    
    \begin{figure}
        \centering
        \includegraphics[width=0.9\columnwidth]{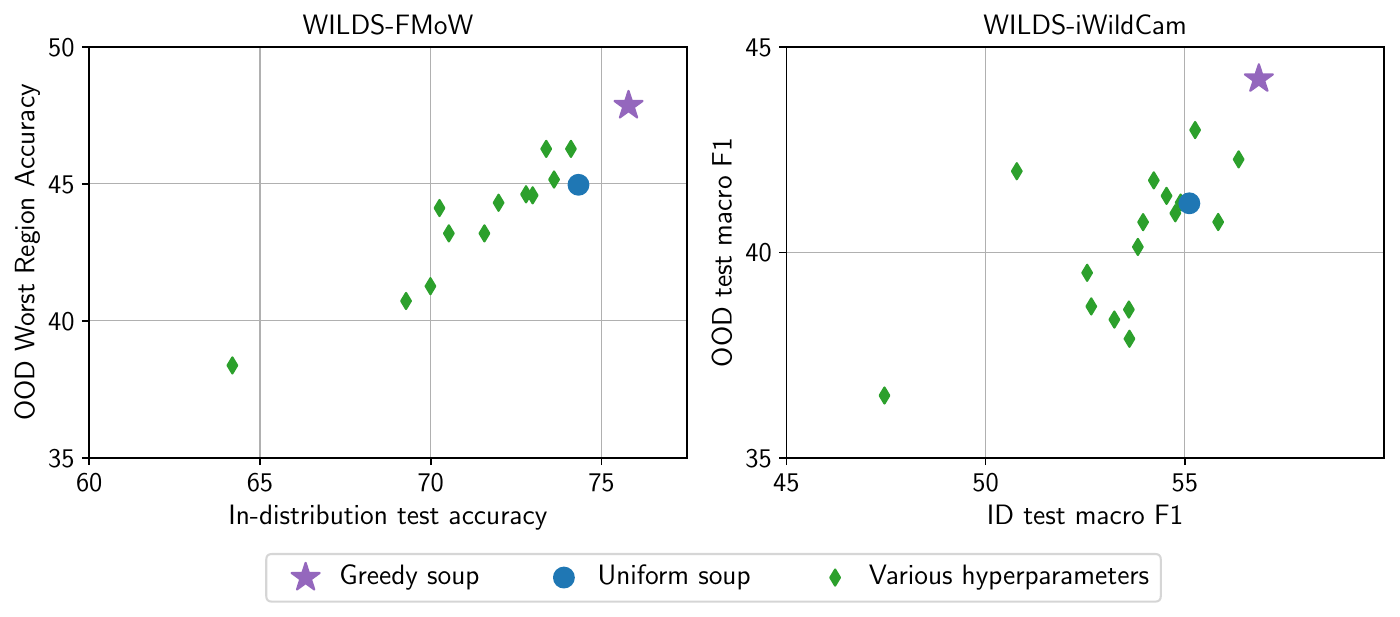}
        \caption{Model soups improve accuracy when fine-tuning on the diverse classification tasks
        WILDS-FMoW~\cite{wilds2021,christie2018functional} and WILDS-iWildCam~\cite{wilds2021,beery2021iwildcam}.
        Results are shown for the CLIP ViT-L/14 model and a random hyperparameter
        search over learning rate, weight-decay, 
        iterations, data augmentation, mixup, and label smoothing.}
        \label{fig:fmow-iwc}
    \end{figure}

    \begin{figure}
            \begin{minipage}[t]{0.48\linewidth}
                \centering
        \includegraphics[width=\textwidth]{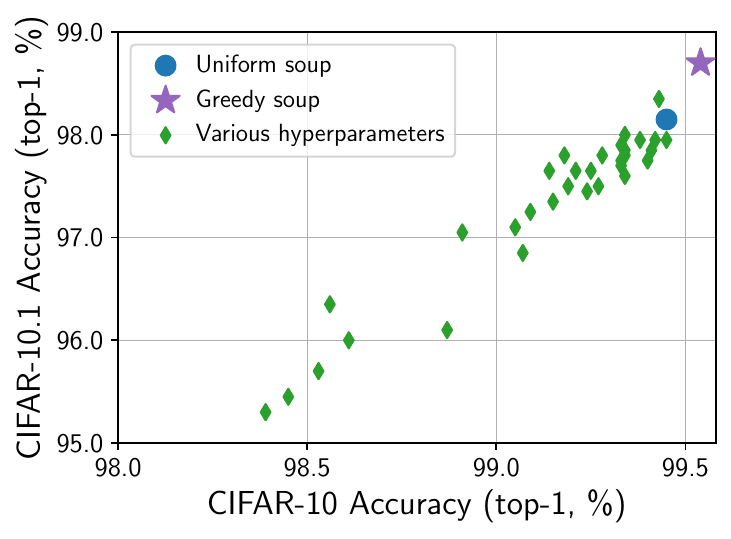}
        \captionof{figure}{Fine-tuning a CLIP ViT-L model
        on CIFAR-10~\cite{krizhevsky2009learning} with the random hyperparameter search described in Section~\ref{app:clipdetails}.
        The $y$-axis displays accuracy on CIFAR-10.1~\cite{pmlr-v97-recht19a},
        a reproduction of CIFAR-10 with a distribution shift.}
        \label{fig:vitlcifar}
              \end{minipage}
            \hfill
            \begin{minipage}[t]{0.46\linewidth}
              \centering
        \includegraphics[width=\textwidth]{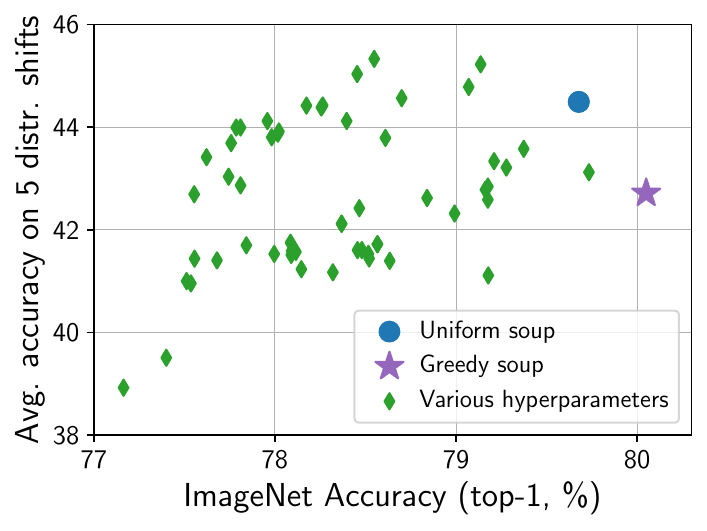}
        \captionof{figure}{Fine-tuning on ImageNet, using a ViT-B/32~\cite{dosovitskiy2021an}
        pre-trained on ImageNet-22k~\cite{deng2009imagenet}.}
        \label{fig:in22k}
            \end{minipage}
          \end{figure}

    In this section we explore fine-tuning or pre-training
    on additional datasets.
    First, Figure~\ref{fig:fmow-iwc} displays results when fine-tuning a CLIP ViT-L model on two datasets included in the WILDS~\cite{wilds2021} challenge, FMoW~\cite{christie2018functional} and iWildCam~\cite{beery2021iwildcam}.
    
    Next, Figure~\ref{fig:vitlcifar} displays results for
    fine-tuning a CLIP ViT-L model
    on CIFAR-10~\cite{krizhevsky2009learning}.
    The $y$-axis of Figure~\ref{fig:vitlcifar}
    displays accuracy on CIFAR-10.1~\cite{pmlr-v97-recht19a}, a reproduction
    of CIFAR-10 with a distribution shift.
    The individual models are fine-tuned with the random hyperparameter
    search described in Section~\ref{app:clipdetails}.
    
    In addition, Figure~\ref{fig:in22k} shows results when fine-tuning a ViT-B/32~\cite{dosovitskiy2021an} model pre-trained on ImageNet-22k~\cite{deng2009imagenet} and fine-tuned on ImageNet.
    This differs from many of our other experiments as the dataset used for pre-training
    is smaller and less diverse.
    While the greedy soup offers an improvement,
    the improvement is less substantial than Figure~\ref{fig:teaser}
    which uses the same model and hyperparameter search
    but a different pre-training dataset.
    
\begin{figure*}[h!]
        \centering
        \includegraphics[width=\textwidth]{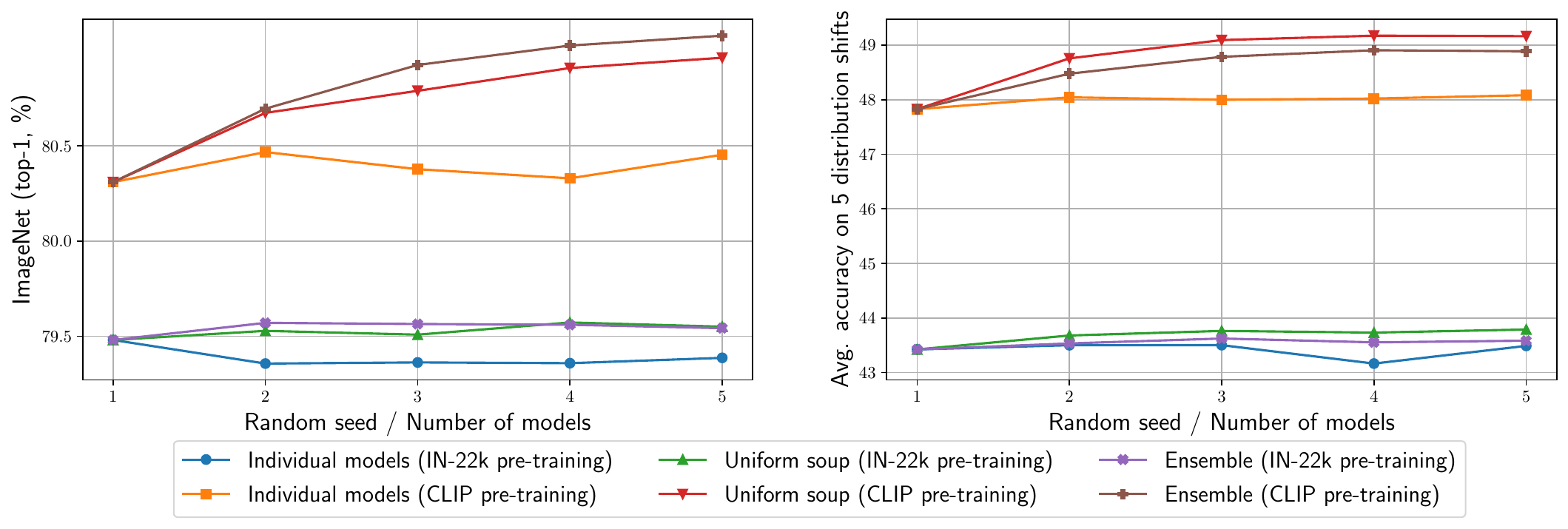}
         \caption{For a CLIP and ImageNet-22k pre-trained ViT-B/32 model,
        we use the best hyperparameters found by the hyperparameter sweep to fine-tune
        multiple times, varying only the random seed.
        For an experimental budget of $1 \le k \le 5$ models, we show:
        (i) the individual model with random seed $k$,
        (ii) the model soup composed of models with random seeds 1 through $k$, and
        (iii) the ensemble composed of models with random seeds 1 through $k$.}
        \label{fig:in22kanalysis}
    \end{figure*}

    Finally, we fine-tune a ViT-B/32 model five times on ImageNet, using the best hyperparameters found by the hyperparameter sweep, varying only the random seed. This experiment is conducted both for a model pre-trained on ImageNet-22k~\cite{deng2009imagenet} and a pre-trained CLIP model.
    The results are shown in Figure~\ref{fig:in22kanalysis}, comparing, for an experimental budget of $1 \le k \le 5$ models:
    (i) the individual model with random seed $k$,
    (ii) the model soup composed of models with random seeds 1 through $k$, and
    (iii) the ensemble composed of models with random seeds 1 through $k$.
    The performance of the model soup appears correlated with the performance
    of the ensemble. Moreover, we find that CLIP models are more amenable to both ensembling and souping than models pre-trained on ImageNet-22k.

    \FloatBarrier
    \section{Additional grid searches and initializations}\label{app:moreinits}
    
    This section recreates Figure~\ref{fig:active} with different initializations (linear probe initialization or zero-shot)
    and different grid searches (standard and extreme grid) when fine-tuning CLIP ViT-B/32.
    The standard and extreme grid searches are described in Section~\ref{app:clipdetails}.

\begin{figure*}[h!]
        \centering
        \includegraphics[width=\textwidth]{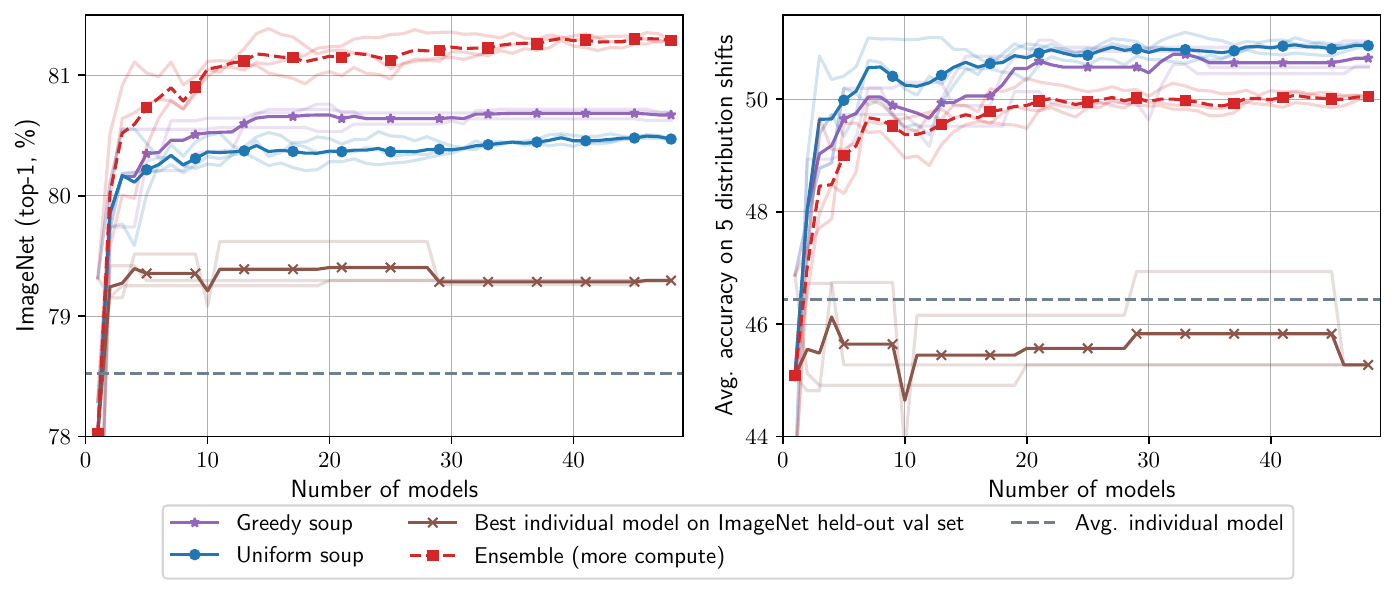}
        \caption{Replicating Figure~\ref{fig:active} with the LP initialization and the
        \emph{standard grid} hyperparameter search.}
        \label{fig:nd61}
    \end{figure*}

\begin{figure*}[h!]
        \centering
        \includegraphics[width=\textwidth]{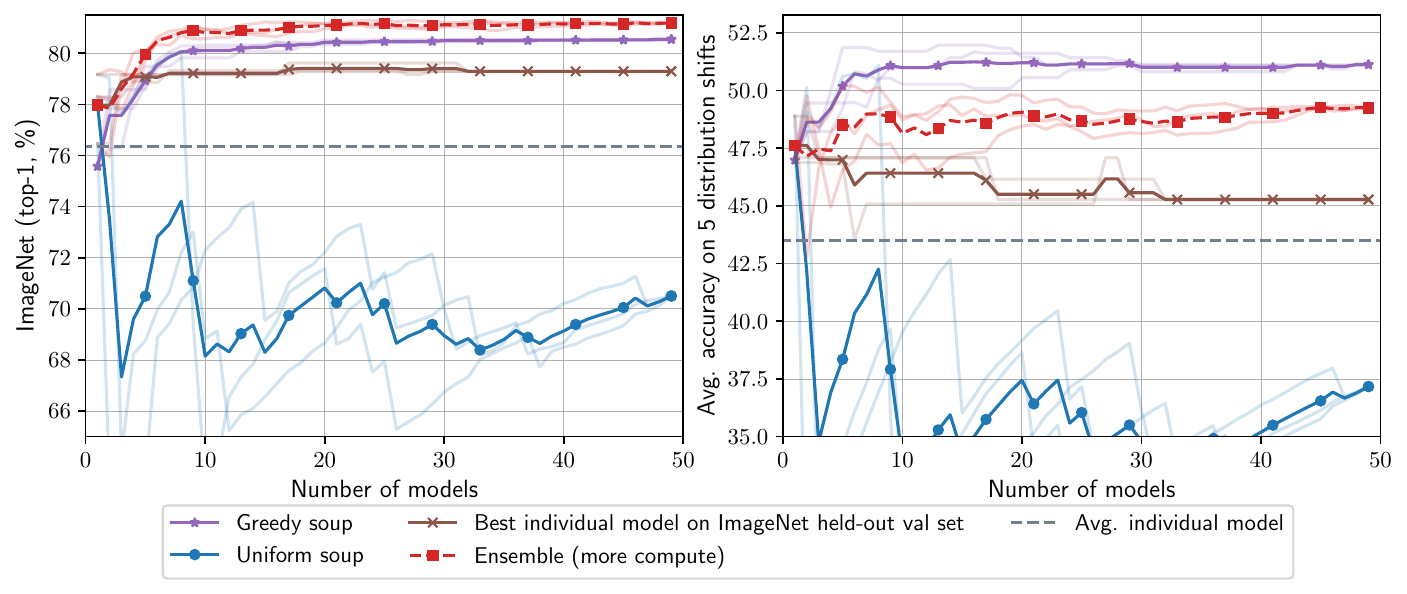}
        \caption{Replicating Figure~\ref{fig:active} with the LP initialization and the
        \emph{extreme grid} hyperparameter search.}
        \label{fig:nd62}
    \end{figure*}

\begin{figure*}[h!]
        \centering
        \includegraphics[width=\textwidth]{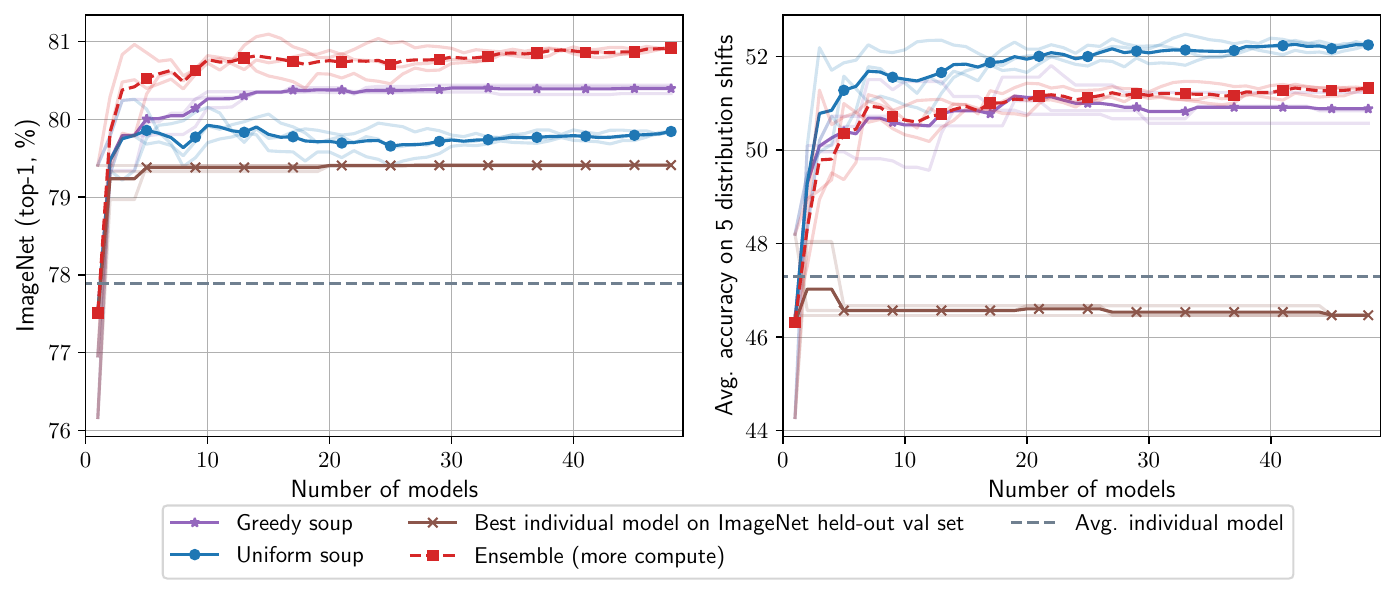}
        \caption{Replicating Figure~\ref{fig:active} with the zero-shot initialization and the
        \emph{standard grid} hyperparameter search.}
        \label{fig:nd11}
    \end{figure*}

\begin{figure*}[h!]
        \centering
        \includegraphics[width=\textwidth]{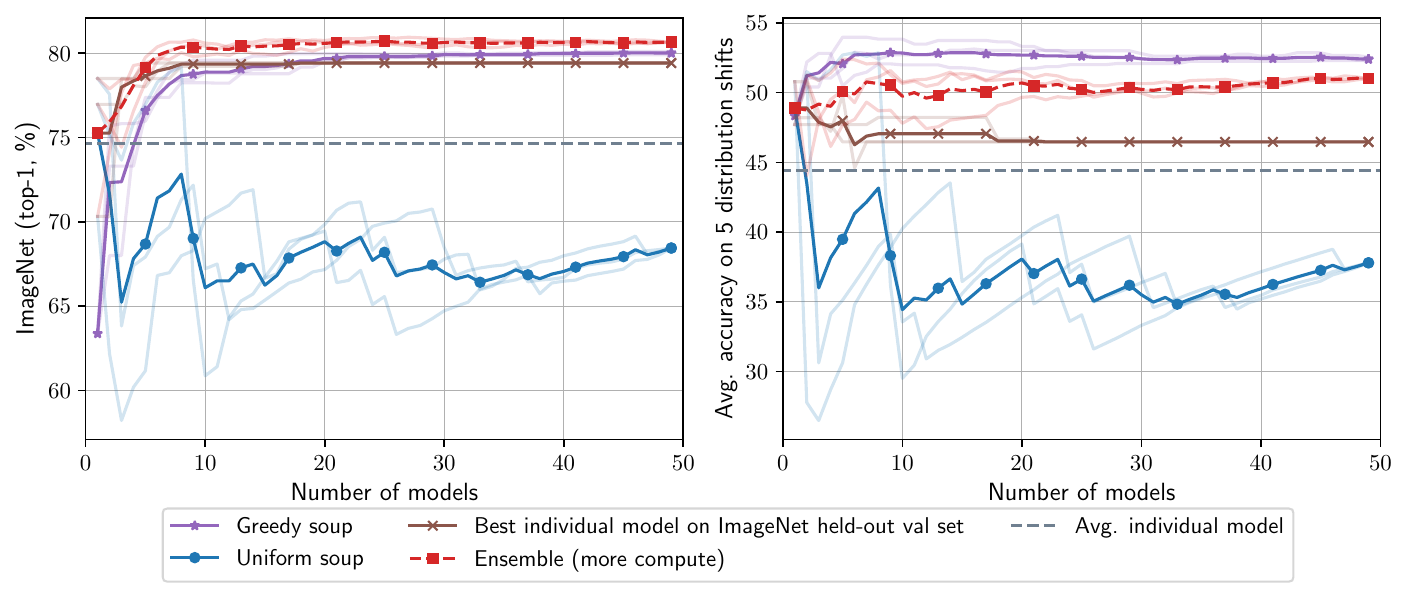}
        \caption{Replicating Figure~\ref{fig:active} with the zero-shot initialization and the
        \emph{extreme grid} hyperparameter search.}
        \label{fig:nd12}
    \end{figure*}

    Figure~\ref{fig:nd61} considers the linear probe (LP) initialization and 
    the \emph{standard grid}.
    Figure~\ref{fig:nd62} considers the linear probe (LP) initialization and 
    the \emph{extreme grid}.
    Figure~\ref{fig:nd11} considers the zero-shot initialization and 
    the \emph{standard grid}.
    Figure~\ref{fig:nd12} considers the zero-shot initialization and 
    the \emph{extreme grid}.


    \FloatBarrier
    \section{Learned soup}\label{app:learned-soup}
    
    In addition to the greedy soup method described in the text, we also explore a more advanced souping procedure, which removes the sequential constraint 
    from the greedy soup and requires only a single pass through the held out validation set.
    We refer to this method as the \emph{learned soup}, as it involves learning the soup mixing coefficients
    for each of the ingredients on the held-out validation set. However, the learned soup has the downside of
    requiring all models to be simultaneously loaded in memory. In practice we combine the models on
    CPU before moving the parameters to GPU for each batch.
    For loss $\ell$ and validation set $\{(x_i, y_i)\}_{i=1}^n$, we find mixing coefficients 
    $\alpha \in \mathbb{R}^k$ and temperature scaling parameter $\beta$ via
    \begin{equation}%
            \argmin_{\alpha \in \mathbb{R}^k, \beta \in \mathbb{R} } \
            \sum_{j=1}^n \ell\mleft(
            \beta \cdot f\mleft(x_j, \ \sum_{i=1}^k {\alpha}_i \theta_i \mright), \ y_j \mright).
    \end{equation}
    In practice we find better results when $\alpha$ is parameterized as the output of a softmax,
    so that each $\alpha_i$ is positive and values sum to one.
    We optimizer the aforementioned equation with gradient based mini-batch optimization for three epochs over
    the held-out validation set with the AdamW otpimizer and constant learning rate 0.1.
    
    As presented in Table~\ref{tab:results}, we also try a ``by layer'' variant
    of the learned soup. For this we learn a separate $\alpha$ for each layer of the network. Finally, another way to get non-uniform mixing coefficients is to sample with replacement in the greedy soup procedure.

    \section{Experimental details}\label{app:details}

    \subsection{Error landscape visualizations}\label{app:error}
    
\begin{figure}
        \centering
        \includegraphics[width=\textwidth]{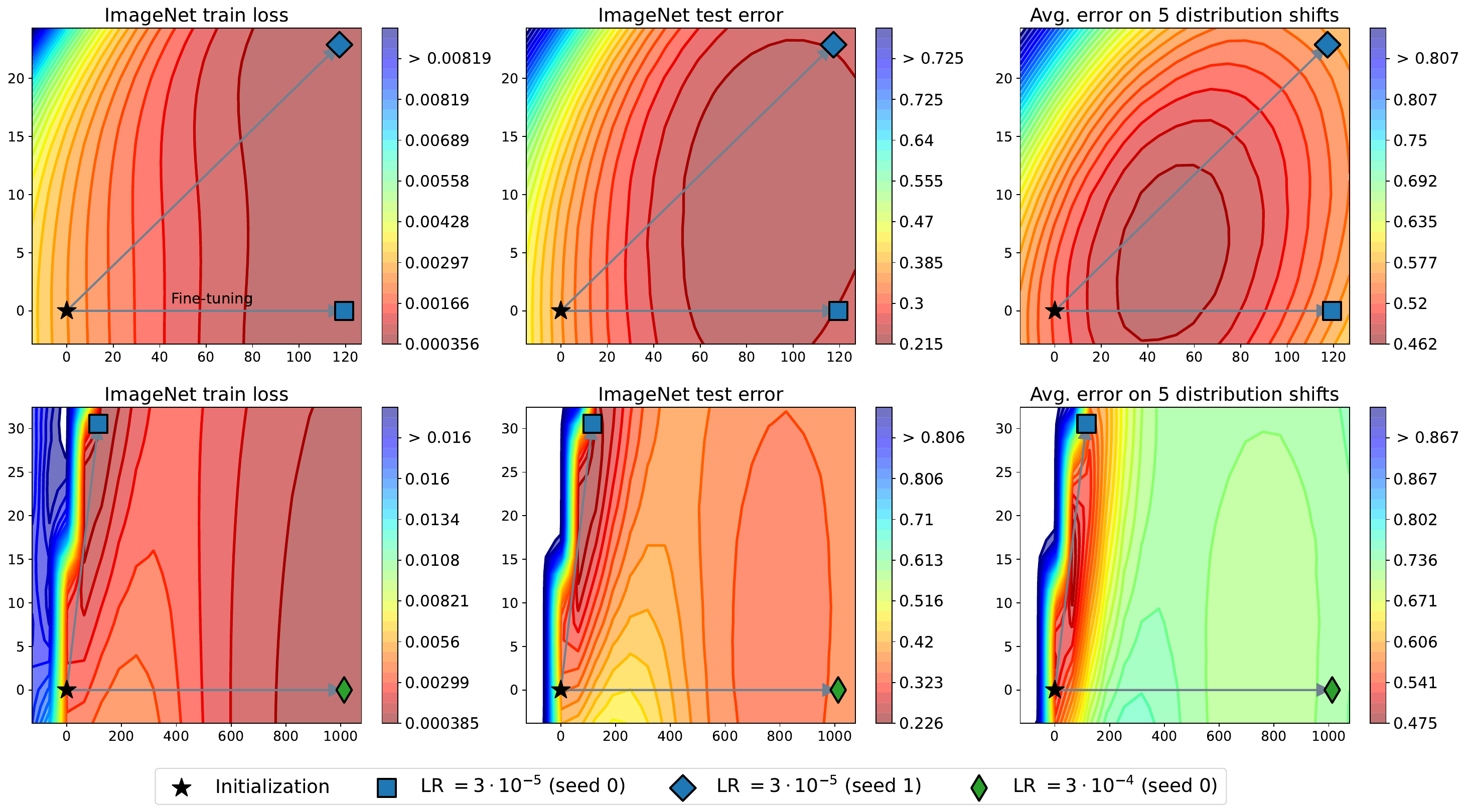}
        \caption{Replicating Figure~\ref{fig:error} with a 10x larger learning rate instead of 10x smaller in the second row.}
        \label{fig:errorbig}
    \end{figure}

    To supplement Figure~\ref{fig:error}, we provide an identical experiment but with a
    10x bigger learning rate instead of 10x smaller. Results are 
    illustrated in Figure~\ref{fig:errorbig} with linear instead of log scaling
    for the contour lines. Since the error difference is more substantial,
    linear scaling was more clear.
    When fine-tuning with a larger learning rate,
    error increases on the path between the two fine-tuned solutions.
    All error landscape visualizations use CLIP ViT-B/32 fine-tuned on ImageNet for 10 epochs
    with minimal data augmentation, as used by CLIP during pre-training.
    When computing angles between the two fine-tuned solutions, as in Figure~\ref{fig:angles},
    we use the repeated weights which constitute the majority of the network parameters.
    We ignore gain terms which tend to skew positive if occurring before ReLU activations.
    
    In Figure~\ref{fig:angles} we consider solutions fine-tuned with learning rates less that $10^{-4}$.
    As in Figure~\ref{fig:errorbig}, if a learning rate that is large is used accuracy will decrease on
    the path in weight space between the two models.
    
    \subsection{Model soups}\label{app:mainres}
    
    This section describes the set of hyperparameters used for searches. For all ImageNet experiments, we withhold 2\% of the training set and use these examples as the held-out validation set for model selection in greedy and learned soup.
    
    \subsubsection{CLIP experiments}\label{app:clipdetails}
    
    Unless otherwise mentioned, all experiments used the AdamW optimizer~\cite{loshchilov2018decoupled}
    with cosine annealing learning rate schedule~\cite{loshchilov2016sgdr} for 10 epochs at batch size 512 at a resolution of 224$\times$224.
    When necessary we discretize augmentation strength into minimal, medium, and strong.
    Minimal augmentation uses only a random crop consisting of 90\%-100\% of the total image area.
    Medium is the default augmentation used by the timm library \cite{rw2019timm}.
    Strong refers to 
    RandAugment~\cite{cubuk2020randaugment} ($N = 2$, $M = 15$).
    
    We now provide the low level details for the hyperparemter searches, which are
    \emph{standard grid}, \emph{extreme grid}, and \emph{random search}.
    The \emph{standard grid} includes learning rates $3 \cdot 10^{-5}, 2 \cdot 10^{-5}, 1 \cdot 10^{-5}, 3 \cdot 10^{-6}$, where $2 \cdot 10^{-5}, 1 \cdot 10^{-5}$ typically perform the best.
    Augmentation strengths are minimal, medium, or strong. Mixup is either off or on at $\alpha = 0.5$.
    We consider all combinations of the above, running each
    hyperparameter configuration with two random seeds.

    The \emph{extreme grid} considers 
    learning rates $3 \cdot 10^{-4}, 1 \cdot 10^{-4}, 3 \cdot 10^{-5}, 2 \cdot 10^{-5}, 1 \cdot 10^{-5}, 3 \cdot 10^{-6}, 1\cdot 10^{-6}, 1\cdot 10^{-7}$, where $2 \cdot 10^{-5}, 1 \cdot 10^{-5}$ typically perform the best.
    Augmentation strengths are minimal, medium, or strong. Mixup is either off or on at $\alpha = 0.5$.
    Moreover, we include the initialization in this search, which often outperforms some of the extreme learning rates but is far from the most accurate model.

    The \emph{random search} chooses learning rate $10^{-\lambda_1}$ where
    $\lambda_1$ is selected uniformly at random  from 4 to 6.
    Weight decay is chosen randomly as $10^{-\lambda_2}$ where
    $\lambda_2$ is selected uniformly at random  from 0.2 to 4.
    With probability 0.5, label smoothing is set to 0 and otherwise it
    is selected uniformly at random between 0 and 0.25.
    Fine-tuning epochs are chosen randomly between four and sixteen.
    Mixup is 0 with probability 0.5, and otherwise is chosen
    uniformly at random from 0 to 0.9.
    With probability $1/3$ we use minimal augmentation,
    otherwise we use randaug where $M$ and $N$ are chosen
    uniformly at random between 0 and 20 and 0 and 2 respectively.
    
    When fine-tuning on WILDS-FMoW and WILDS-iWildCam for Figure~\ref{fig:fmow-iwc},
    we use the same random search as when we fine-tune CLIP on ImageNet.
    The only difference is that we are able to use a larger ViT-L/14 model
    as the datasets are smaller. This also requires us to change the default batch size
    from 512 to 128.
    
    \subsubsection{ALIGN experiments} \label{app:aligndetails}
    We fine-tuned ALIGN EfficientNet-L2 models using AdamW with weight decay of 0.1 at a resolution of $289 \times 289$ for 25 epochs, with the final layer initialized from a linear probe without data augmentation. We fine-tuned 5 models with standard Inception-style random crops (consisting of 5\% to 100\% of the total image area with an aspect ratio between 0.75 and 1.33) and different learning rates ($1 \cdot 10^{-6}$, $2 \cdot 10^{-6}$, $5 \cdot 10^{-6}$, $1 \cdot 10^{-5}$, and $2 \cdot 10^{-5}$). We also fine-tuned 7 additional models at a learning rate of $5 \cdot 10^{-6}$ with different data augmentation strategies. Specifically, we varied the random cropping strategy (either Inception-style crops or less aggressive crops consisting of 90\% to 100\% of the total image area with an aspect ratio between 0.95 and 1.05), the use of RandAugment~\cite{cubuk2020randaugment} (off or $N = 2$, $M = 15$), and the use of mixup~\cite{zhang2017mixup} (off or $\alpha = 0.5$) and trained models with all combinations of these strategies. Our soups are obtained by considering these 12 models as well as the linear probe initialization. We perform evaluation at $360 \times 360$ resolution using a square center crop from images. The accuracy we attain with greedy soup approaches that reported by \citet{jia2021scaling}, which evaluated at $600 \times 600$ resolution.
    
    \subsubsection{ViT-G/14 experiments}
    \label{app:vitgdetails}
    These models are initialized with a backbone that was pretrained on the JFT-3B dataset~\cite{zhai2021scaling} and linear probes obtained at either the $224\times 224$ resolution at which the ViT-G/14 was pretrained or at the $518\times 518$ resolution used for fine-tuning. Models are fine-tuned at a batch size of 512 for either 10,000 or 20,000 steps (approximately 4 or 8 epochs) using the Adafactor optimizer~\cite{shazeer2018adafactor} with learning rates of $3 \cdot 10^{-5}$ or $5 \cdot 10^{-5}$; a constant or cosine decay learning rate schedule; and softmax or binary cross-entropy loss. When fine-tuning with binary cross-entropy loss, we use a linear probe that is also trained with binary cross-entropy loss. We vary data augmentation, applying RandAugment~\cite{cubuk2020randaugment}, mixup~\cite{zhang2017mixup}, or CutMix~\cite{yun2019cutmix} of varying strengths and random cropping with a minimum crop size of 5\%, 70\%, 90\%, or 100\% of the full image.  When applying SAM, we consider models with perturbations either synchronized or unsynchronized across accelerators, including one model with synchronized perturbations and a combination of CutMix and SAM. All models are fine-tuned at $518 \times 518$ resolution and evaluated by rescaling test images to $550 \times 550$ (without preserving the aspect ratio) and taking a $518 \times 518$ central crop.
    
    We manually tuned hyperparameters with the goal of maximizing single-model accuracy. After settling on the use of Adafactor as the optimizer, we included all subsequently trained models in the pool of models to be used for greedy soup. The model that performs best on the holdout set is initialized with a $224\times 224$ linear probe and fine-tuned with a learning rate of 3e-5 and a constant learning rate decay schedule, with softmax cross-entropy loss, a minimum crop size of 90\%, and CutMix with $\alpha = 0.2$. The model that performs best on the official ImageNet validation set is initialized with a $518\times 518$ linear probe and fine-tuned at a learning rate of 3e-5 and a constant learning rate decay schedule, with softmax cross-entropy loss, a minimum crop size of 90\%, CutMix with $\alpha = 0.2$, and SAM. The greedy soup contains models trained with a wide range of different hyperparameter values including different learning rates, linear probes, loss functions, and every form of data augmentation and minimum crop size investigated. Notably, although models trained with SAM with synchronized perturbations are included in the greedy soup, the greedy soup process skips over the models trained with SAM with unsynchronized perturbations because adding them produces a large drop in holdout accuracy.
    
    \subsection{Cross-dataset soups details}\label{app:crossdataset}
    When fine-tuning we initialize with CLIP ViT-B/32 and use learning rate $3\cdot 10^{-5}$ for 10 epochs with mini-batch size of 512.
    We train with minimal augmentation.
    
    \subsection{Text classification datasets}
    \label{app:nlp_datasets}
    
    We study four text classification datasets from the GLUE benchmark \cite{wang2018glue}.
    
    \paragraph{Microsoft Research Paraphrase Corpus} (MRPC; \cite{dolan2005automatically}) contains pairs of sentences, labeled as either nearly semantically equivalent, or not. The dataset is evaluated using the average of $F_1$ and accuracy. The training set consists of 3.7 thousand samples and the validation set of 409 samples.
    
    \paragraph{Recognizing Textual Entailment} (RTE;   \cite{wang2018glue}) contains pair of sentences, and the task is to predict whether the first sentence (the premise) entails or contradicts the second sentence (the hypothesis). The data  is originally from a series of datasets \cite{dagan2005pascal, bar2006second, giampiccolo2007third, bentivogli2009fifth}. The dataset is evaluated using classification accuracy. The training set consists of 2.5 thousand samples and the validation set of 277 samples.
    
    \paragraph{Corpus of Linguistic Acceptability} (CoLA;
     \cite{warstadt2018neural}) contains sentences labeled as either grammatical or ungrammatical. Models are evaluated on Matthews correlation (MCC; \cite{matthews1975comparison}), which ranges between $-1$ and $1$. The training set consists of 8.6 thousand samples and the validation set consists of 1043 samples.
     
    \paragraph{Stanford Sentiment Treebank} (SST-2; \cite{socher2013recursive}) contains sentences labelled as expressing \textit{positive} or \textit{negative} sentiment, collected from movie reviews. The dataset is evaluated using classification accuracy. The training set consists of 67 thousand samples and the validation set consists of 873 samples.
    \subsection{Fine-tuning details for text classification tasks}
    \label{app:nlp_ft}

        \begin{table}
            \caption{Performance of model soups on four text classification datasets from the GLUE benchmark \cite{wang2018glue}.}
            \begin{center}\small
            \setlength{\arrayrulewidth}{.01em}
            \setlength{\tabcolsep}{12pt}
            \begin{tabular}{llcccc}
                \toprule
                Model & Method & MRPC & RTE & CoLA & SST-2  \\\midrule
                \multirow{3}{*}{BERT-base \cite{devlin2019bert}} & Best individual model & 88.3 & 61.0 & 59.1 & 92.5 \\
                & Uniform soup & 76.0 & 52.7 & 0.0 & 89.9 \\
                & Greedy soup & 88.3 & 61.7 & 59.1 & 93.0 \\\midrule
                \multirow{3}{*}{BERT-large \cite{devlin2019bert}}& Best individual model & 88.8 & 56.7 & \textbf{63.1} & 92.2 \\
                & Uniform soup & 15.8 & 52.7 & 1.90 & 50.8 \\
                &  Greedy soup & 88.8 & 56.7 & \textbf{63.1} & 92.3 \\\midrule
                \multirow{3}{*}{T5-small \cite{raffel2020t5}} & Best individual model & 89.7 & 70.0 & 42.2 & 91.7 \\
                &  Uniform soup & 82.7 & 61.7 & 10.4 & 91.1 \\
                &  Greedy soup & 89.7 & 70.0 & 43.0 & 91.7 \\\midrule
                \multirow{3}{*}{T5-base \cite{raffel2020t5}} & Best individual model & 91.8 & 78.3 & 58.8 & 94.6\\
                &  Uniform soup & 86.4 & 71.8 & 12.3 & 94.6 \\
                &  Greedy soup & 92.4 & 79.1 & 60.2 & 94.7 \\\midrule
                \multirow{3}{*}{T5-large \cite{raffel2020t5}} & Best individual model & \textbf{93.4} & 82.7 & 61.7 & \textbf{96.3} \\
                &  Uniform soup & 74.8 & 50.2 & 0.00 & 96.0 \\
                &  Greedy soup & \textbf{93.4} & \textbf{84.8} & 62.7 & \textbf{96.3} \\
                \bottomrule
            \end{tabular}
            \end{center}
            \label{tab:nlp_full}
        \end{table}

    Each model is fine-tuned 32 times on each dataset, performing a random hyperparameter search. The learning rate is chosen uniformly in log space over [$10^{-6}$, $10^{-3}$], the batch size is chosen uniformly from $\{8, 16, 32, 64\}$ and the number of epochs from $\{2,3,5\}$. Evaluation is conducted once at the end of training, without early stopping. We use a maximum sequence length of 128 tokens and train with Adam \cite{kingma2014adam} using $\beta_1=0.9$, $\beta_2=0.999$ and $\epsilon=10^{-8}$, gradient clipping of $1.0$, no weight decay, and with the learning rate being decayed linearly to zero at the end of training. We use pre-trained weights from the Huggingface Transformers library \cite{wolf2020transformers}. For BERT models, we use the uncased version. 
    
    Fine-tuning occurs without any additional parameters to avoid distorting the features from the pre-trained models \cite{kumar2021finetuning}. For such, the classification tasks are adapted to be suited to the pre-training objective of BERT and T5. For T5, the tasks are cast as a sequence-to-sequence problem. For instance, for sentiment analyses, an example is to predict ``\texttt{A) positive}" from ``\texttt{sentence: The best movie I've ever seen! | options: A) positive B) negative | label:}". For BERT, the tasks are cast as a masked language modeling problem. For instance, for linguistic acceptability, an example is to predict ``\texttt{A) acceptable}" for the inputs ``\texttt{sentence: model soups are grammatical. | options: A) acceptable B) unacceptable | label: [MASK] [MASK] [MASK]}". For evaluation, we select which of the options is given the highest probability according to the model.
    
    The full set of results is shown in Table \ref{tab:nlp_full}.  On 10 out of the 20 combinations of models and datasets, the greedy soup shows better performance than the best individual model from the hyperparameter search. Uniform soups show worse performance than the best individual model on all experiments, which could be an artifact of the broad range of hyperparameters used in the search. While the experiments varied only basic hyperparameters such as learning rate and batch size, we hypothesize that a broader set of hyperparameter choices (e.g. data augmentation \cite{wei2019eda,ma2019nlpaug}) could lead to more diverse models and better soups.
    
    Finally, as a word of caution for practitioners, we remind readers that many recent language models have tied weights on the output and embedding layers \cite{press2017using}. For this reason, caution is needed when writing code to average models in-place.
    
    \FloatBarrier

    \FloatBarrier
    \section{Analytical comparison details}\label{app:theory}

    \newcommand{\ones}{\mathbf{1}}
\newcommand{\sbv}[1]{e^{(#1)}}

\subsection{Notation and preliminaries}
We begin by restating and adding to the notation used in Section~\ref{sec:theory}.
For a model with parameter vector $\theta\in\R^d$ and input vector $x$, we let $f(x;\theta)\in\R^C$ denote the model's logit output for $C$-way classification. Throughout, we fix two endpoint models $\theta_0$ and $\theta_1$, and for an interpolation parameter $\alpha\in[0,1]$ define
\begin{equation*}
	\theta_\alpha \defeq (1-\alpha) \theta_0 + \alpha \theta_1,
	~~\mbox{and}~~
	\fwse(x) \defeq f(x; \theta_\alpha)
\end{equation*}
to be the ``soup'' weight averaged model and its corresponding logits. We also write
\begin{equation*}
	\fose(x) \defeq (1-\alpha)f(x;\theta_0) + \alpha f(x;\theta_1)
\end{equation*}
for the logits of the ensemble model. We write
\begin{equation*}
	\delta = \theta_1 - \theta_0
\end{equation*}
for the difference of the two endpoints.

For a logit vector $f\in\R^C$ and a ground-truth label $y$, denote the cross-entropy loss by
\begin{equation*}
	\ell(f;y) = \log\left(\sum_{y'}\exp\{f_{y'}-f_{y}\}\right).
\end{equation*}
For some distribution over $x,y$ we write the expected $\beta$-calibrated log losses of the soup and ensemble as
\begin{equation*}
	\exloss_\alpha = \E_{x,y} \ell(\beta f(x;\theta_\alpha),y)
	~~\mbox{and}~~
	\exlossens_\alpha = \E_{x,y} \ell(\beta \fose(x),y),
\end{equation*}
respectively.

We have the following expression for the derivatives of cross entropy w.r.t.\ logits. The gradient is
\[
\grad_{f}\ell\left(f,y\right)=\softmax(f) - \sbv{y},
\]
where $\sbv{i}$ is the $i$th standard basis vector and $\softmax(f)\in\R^C$ has $e^{f_i}/\sum_j e^{f_j}$ in its $i$th entry. The Hessian is
\[
\hess_{f}\ell\left(f,y\right)=\mathrm{diag}\left(\softmax\left(f\right)\right)
-
[\softmax(f)] [\softmax(f)]^T,
\]
so that for any $v\in \R^C$, we have
\begin{equation*}
	v^T \hess_{f}\ell\left(f,y\right) v = \mathrm{Var}_{Y\sim \softmax(f)} [v_Y].
\end{equation*}

Finally, we use $\delta^T \grad f(x;\theta)$ to denote a vector in $\R^C$ whose $i$th entry is $\delta^T \grad [f(x;\theta)]_i$. Similarly, 
 $\delta^T \hess f(x;\theta) \delta$  denotes a vector in $\R^C$ whose $i$th entry is $\delta^T [\hess f(x;\theta)]_i \delta$, where gradients and Hessian are with respect to $\theta$.

\subsection{An exact expression for logit difference}
We use the fundamental theorem of calculus and elemntary algebraic manipulation to obtain an exact integral form for the difference between the soup and ensemble logits. To streamline notation we drop the dependence of the logits on the input $x$.

\begin{align}
	\fose-\fwse & =\left(1-\alpha\right)\left[f\left(\theta_{0}\right)-f\left(\theta_{\alpha}\right)\right]+\alpha\left[f\left(\theta_{1}\right)-f\left(\theta_{\alpha}\right)\right]\nonumber \\
	& =-\left(1-\alpha\right)\int_{0}^{\alpha}\delta^{T}\grad f\left(\theta_{t}\right)\d t+\alpha\int_{\alpha}^{1}\delta^{T}\grad f\left(\theta_{t}\right)\d t\nonumber \\
	& =-\left(1-\alpha\right)\int_{0}^{\alpha}\left[\delta^{T}\grad f\left(\theta_{\alpha}\right)+\int_{\alpha}^{t}\delta^{T}\grad f\left(\theta_{\tau}\right)\delta \d\tau\right]\d t+\alpha\int_{\alpha}^{1}\left[\delta^{T}\grad f\left(\theta_{\alpha}\right)+\int_{\alpha}^{t}\delta^{T}\grad f\left(\theta_{\tau}\right)\delta \d\tau\right]\d t\nonumber \\
	& =-\left(1-\alpha\right)\int_{0}^{\alpha}\int_{\alpha}^{t}\left(\delta^{T}\hess f\left(\theta_{\tau}\right)\delta\right)\d\tau \d t+\alpha\int_{\alpha}^{1}\int_{\alpha}^{t}\left(\delta^{T}\hess f\left(\theta_{\tau}\right)\delta\right)\d\tau \d t\nonumber \\
	& =\left(1-\alpha\right)\int_{0}^{\alpha}\int_{t}^{\alpha}\left(\delta^{T}\hess f\left(\theta_{\tau}\right)\delta\right)\d\tau \d t+\alpha\int_{\alpha}^{1}\int_{\alpha}^{t}\left(\delta^{T}\hess f\left(\theta_{\tau}\right)\delta\right)\d\tau \d t\nonumber \\
	& =\left(1-\alpha\right)\int_{0}^{\alpha}\left(\delta^{T}\hess f\left(\theta_{\tau}\right)\delta\right)\d\tau\int_{0}^{\tau}\d t+\alpha\int_{\alpha}^{1}\left(\delta^{T}\hess f\left(\theta_{\tau}\right)\delta\right)\d\tau\int_{\tau}^{1}\d t\nonumber \\
	& =\int_{0}^{1}\left(\delta^{T}\hess f\left(\theta_{\tau}\right)\delta\right)w_{\alpha}\left(\tau\right)\d\tau,
	\label{eq:fose-fwse}
\end{align}
where
\[
w_{\alpha}\left(\tau\right)=\begin{cases}
	\left(1-\alpha\right)\tau & \tau\le\alpha\\
	\alpha\left(1-\tau\right) & \text{otherwise}
\end{cases}=\min\left\{ \left(1-\alpha\right)\tau,\alpha\left(1-\tau\right)\right\}.
\]
Note that
\begin{equation*}
	\int_0^1 w_\alpha(\tau)\d \tau = \frac{\alpha (1-\alpha)}{2}.
\end{equation*}

\subsection{Derivation of approximation}\label{app:theory-deriv}

We continue to suppress the dependence on $x$ in order to simplify notation.
We begin with the following first order approximation of the pointwise log-loss difference between the ensemble and soup, which is also a lower bound due to convexity. 
\begin{equation*}
	\ell(\fose;y) - \ell(\fwse;y)
	\approx
	[\grad_f \ell (\fose; y)]^T (\fose - \fwse) + O\left( ( (\fose - \fwse)^2 ) \right).
\end{equation*}

Now, we approximate the ensemble and soup logit difference using eq.~\ref{eq:fose-fwse} by assuming that $\delta^{T}\hess f\left(\theta_{\tau}\right)\delta \approx \delta^{T}\hess f\left(\theta_{\alpha}\right)\delta$ for all $\tau\in[0,1]$; this holds when the logits are approximately quadratic along the line between the checkpoints. The resulting approximation is
\begin{equation*}
	\fose - \fwse \approx c_\alpha\, \delta^{T}\hess f\left(\theta_{\alpha}\right)\delta
	+
	O\left( \max_{\tau\in[0,1]} | \grad^3 f(\theta_\tau)[\delta^{\otimes 3}] | \right)
	,~~\mbox{where}~c_\alpha \defeq \frac{(1-\alpha)\alpha}{2}.
\end{equation*}

Combining the two approximation above, we obtain
\begin{equation*}
	\ell(\fose;y) - \ell(\fwse;y) \approx c_\alpha\, [\grad_f \ell (\fose; y)]^T \delta^{T}\hess f\left(\theta_{\alpha}\right)\delta.
\end{equation*}

To relate this expression to the Hessian of the loss with respect to the parameters, we note that for any $\theta$ (by the chain rule)
\begin{equation*}
	\delta^T  \hess_\theta \ell( f(\theta); y) \delta 
	=
	[\delta^T \grad f(\theta)]^T \hess_f \ell(f(\theta);y) [\delta^T \grad f(\theta)]
	+
	\grad_f \ell (f(\theta); y)]^T \delta^{T}\hess f(\theta)\delta.
\end{equation*} 
When setting $\theta=\theta_\alpha$, we note that the second term on the RHS is (up to a constant) our approximation for the loss difference). Recalling the expression for the cross-entropy Hessian, the first term is
\begin{equation*}
	[\delta^T \grad f(\theta)]^T \hess_f \ell(f(\theta);y) [\delta^T \grad f(\theta)]
	=
	\mathrm{Var}_{Y\sim \softmax(f(\theta))} \left[ \delta^T \grad f(\theta) \right].
\end{equation*}

As a final approximation, we let
\begin{equation*}
	\delta^T \grad f(\theta_\alpha) \approx f(\theta_1) - f(\theta_0) 
	+
	O\left( \delta^{T}\hess f(\theta)\delta \right);
\end{equation*}
this holds when logits are too far from linear in $\theta$. 

Substituting back and making $x$ explicit, we obtain
\begin{equation*}
	\ell(\fwse(x);y) - \ell(\fose(x);y) \approx
	 -c_\alpha\, \frac{\d^2}{\d\alpha^2} \ell( \fwse(x); y)
	 +c_\alpha\, \mathrm{Var}_{Y\sim \softmax(\fwse(x))} \left[ f(x;\theta_1) - f(x;\theta_0) \right],
\end{equation*}
where we have used
\begin{equation*}
	\delta^T  \hess_\theta \ell( \fwse(x); y) \delta = \frac{\d^2}{\d\alpha^2} \ell( \fwse(x); y). 
\end{equation*}

Scaling all logits by $\beta$, the approximation becomes
\begin{equation*}
	\ell(\beta\fwse(x);y) - \ell(\beta\fose(x);y) \approx
	-c_\alpha\, \frac{\d^2}{\d\alpha^2} \ell( \beta \fwse(x); y)
	+c_\alpha\, \beta^2\mathrm{Var}_{Y\sim \softmax(\beta\fwse(x))} \left[ f(x;\theta_1) - f(x;\theta_0) \right].
\end{equation*}
Averaging the result over $x$, we arrive at the approximation~\eqref{eq:approx}, which we repeat here for ease of reference:
\begin{equation*}
	\exloss_\alpha - \exlossens_\alpha \approx
	-c_\alpha\, \frac{\d^2}{\d\alpha^2} \exloss_\alpha
	+c_\alpha\, \beta^2 \E_x \mathrm{Var}_{Y\sim \softmax(\beta\fwse(x))} \left[ f(x;\theta_1) - f(x;\theta_0) \right].
\end{equation*}

\subsection{Detailed empirical evaluations}\label{app:theory-eval}

\paragraph{Evaluation setup.} We evaluated our bounds on checkpoints from the ViT-B/32 fine-tuning experiments from the \emph{extreme grid} search described in Section~\ref{app:clipdetails}. We selected three learning rate values ($10^{-6}$, $10^{-5}$ and $10^{-4}$), two levels augmentation (none and RandAugment+MixUp), and considered two different random seeds ($0$ and $1$). From these checkpoints (as well as the initialization) we constructed the following $(\theta_0, \theta_1)$ pairs:
\begin{itemize}
	\item All pairs with different learning rate, the same augmentation level and seed 0,
	\item All pairs with the same learning rate, different augmentation level and seed 0,
	\item All pairs with the same learning rate and augmentation level, but different seeds,
	\item All checkpoints with seed 0 coupled with the initialization.
\end{itemize}  
This results in 21 pairs overall. For each pair and each $\alpha\in\{0,0.1,\ldots,0.9,1.0\}$ we evaluated $\err_\alpha, \errens_\alpha, \exloss_\alpha, \exlossens_\alpha$, as well as the approximation~\eqref{eq:approx}. We performed this evaluation on the ImageNet validation set as well as on the 5 OOD test sets considered throughout this paper.  

\paragraph{The effect of temperature calibration.}
Since our ultimate goal is to accurately predict the difference in error rather than the difference in loss, we introduce the inverse-temperature parameter $\beta$ to the loss, and tune it to calibrate the soup model. Specifically, for every model pair, value of $\alpha$ and test set, we take $\beta = \argmin_\beta \E_{x,y} \ell(\beta \fwse(x); y)$. 

\begin{figure*}
        \centering
        \includegraphics[width=\textwidth]{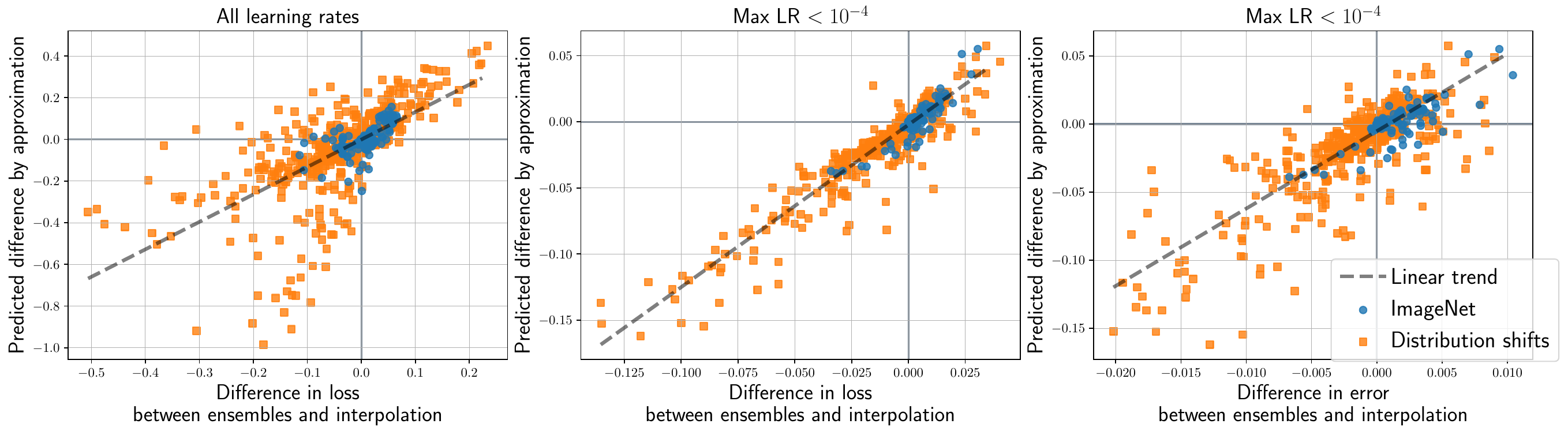}
        \includegraphics[width=\textwidth]{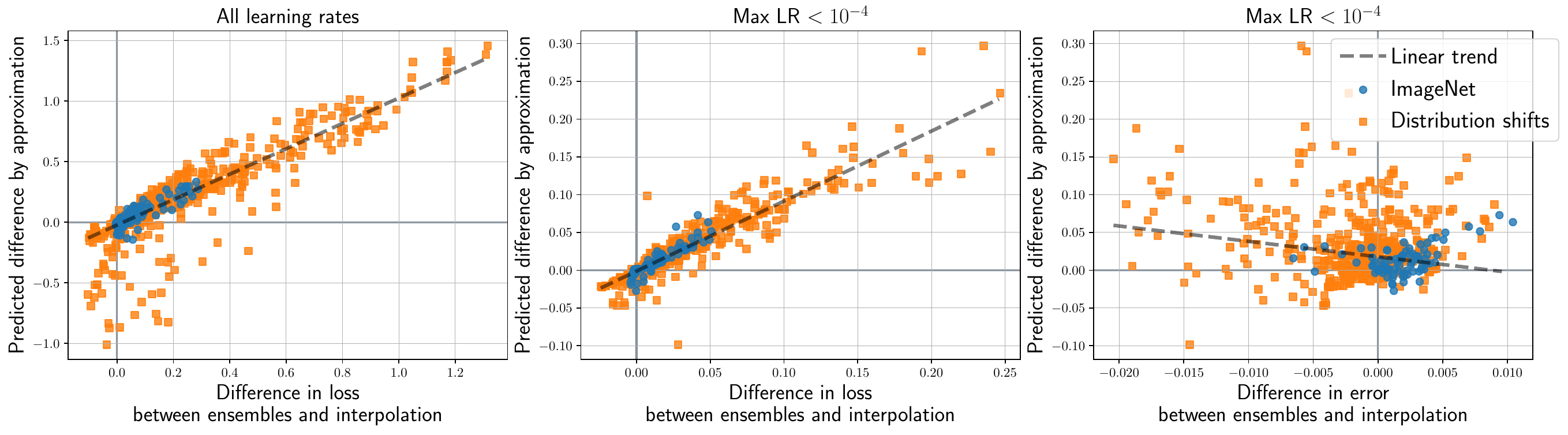}
        \caption{Validation of the analytical approximation~\eqref{eq:approx} for the performance difference of a 2-model soup and ensemble. Each marker on the scatter plots represent a different choice of endpoint models ($\theta_0,\theta_1$) and interpolation weight $\alpha$. In every scatter plot, the vertical axis shows the true performance difference between the soup and ensemble (in loss for the \textbf{left} and \textbf{center} panes, and error for the \textbf{right} pane), where a positive value indicates the ensemble is better. The horizontal axis shows our approximation for the loss difference. The \textbf{top} row shows results with inverse temperature $\beta$ chosen to calibrate the soup, and the \textbf{bottom} row shows results for $\beta$ fixed to $1$.}
        \label{fig:theory-eval}
    \end{figure*}

While choosing $\beta$ based on the soup rather the ensemble might skew the loss in favor of the soup, it has no effect on the difference in prediction error. Moreover, in preliminary experiments calibrating the ensemble produced very similar results. In contrast, as shown in Figure~\ref{fig:theory-eval}, fixing $\beta=1$ throughout results in far poorer prediction of the difference in error.  
    \section{Additional baselines}\label{app:baselines}
    
\begin{figure}
    \centering
    \includegraphics[width=\textwidth]{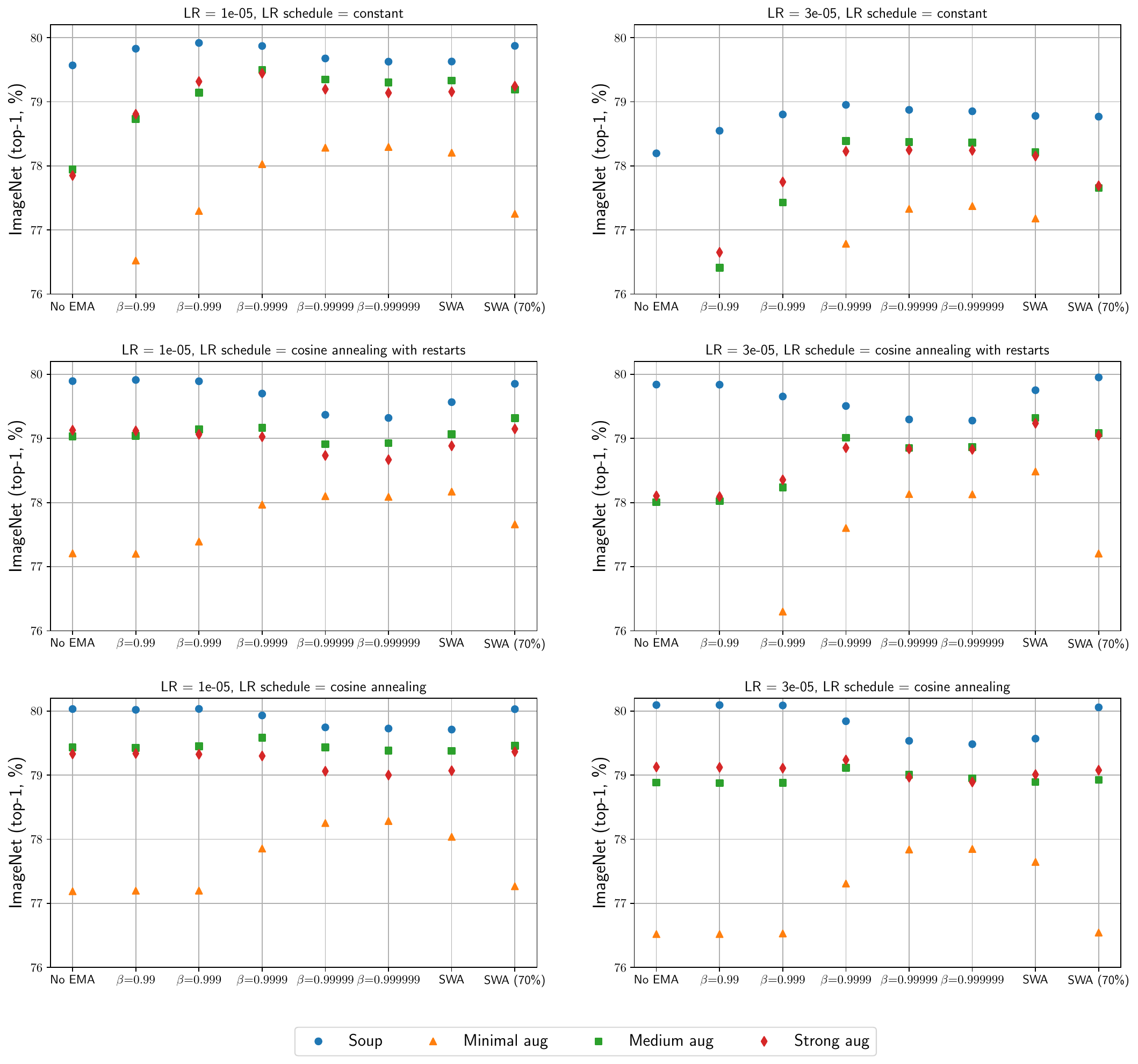}
    \caption{The improvements offered by model soups are additive with weight-averaging along a trajectory (by SWA or EMA with decay $\beta$).
    The soup is the average of the model with minimal, medium and strong data aug.
    Results are shown for a CLIP ViT-B/32 model fine-tuned on ImageNet. For SWA, we average checkpoints which are saved after each of the 10 epochs, while SWA 70\% only averages checkpoints after fine-tune is 70\% complete.}
    \label{fig:swa}
\end{figure}

\begin{figure}
    \centering
    \includegraphics[width=\textwidth]{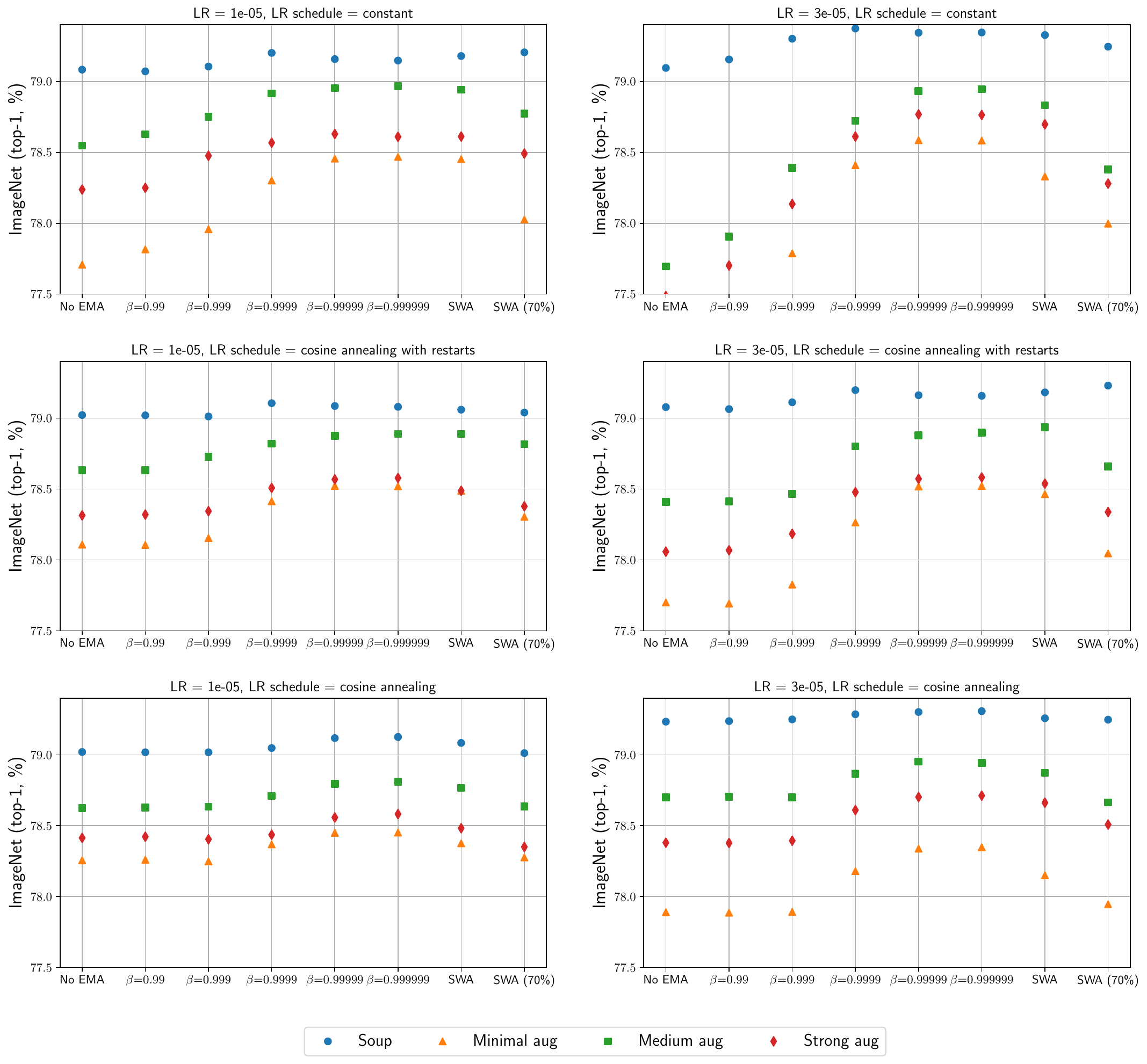}
    \caption{The improvements offered by model soups are additive with weight-averaging along a trajectory (by SWA or EMA with decay $\beta$).
    The soup is the average of the model with minimal, medium and strong data aug.
    Results are shown for a ImageNet-21k pre-trained ViT-B/32 model fine-tuned on ImageNet. For SWA, we average checkpoints which are saved after each of the 10 epochs, while SWA 70\% only averages checkpoints after fine-tune is 70\% complete.}
    \label{fig:swa-in22k}
\end{figure}

    This section explores additional baselines for model soups, including
    distillation from an ensemble as in~\citet{hinton2014dark} (Table~\ref{tab:distill}),
    fix-augmentation as in~\citet{fixaug} (Table~\ref{tab:fixaug}),
    weight-averaging along a trajectory as in~\citet{szegedy2016rethinking, izmailov2018averaging} (Figures~\ref{fig:swa} and \ref{fig:swa-in22k}),
    and Sharpness Aware Minimization as in~\citet{foret2021sharpnessaware} (Table~\ref{tab:sam}).
    
    Unless otherwise mentioned, we fine-tune CLIP ViT-B/32 models
    with AdamW~\cite{loshchilov2018decoupled} and cosine annealing learning rate~\cite{loshchilov2016sgdr}
    for 10 epochs on ImageNet with a learning rate of 2e-5 and medium augmentation (data
    augmentation policies are discussed in more detail in 
    Section~\ref{app:clipdetails}).
    
    We explore the baseline of distillation~\cite{hinton2014dark,hinton2015distilling}
    from the ensemble of three models trained with different
    data augmentation.
    As previously reported~\cite{bagherinezhad2018label,beyer2021knowledge}, we find
    that it improves accuracy to run distillation with data augmentation.
    Unfortunately, this substantially increases the computational resources necessary to distill from the ensemble.
    As we cannot cache the predictions
    of the models in the ensemble, it is necessary to perform a forward pass for each model in the ensemble at each step of fine-tuning.
    This makes distilling from an ensemble
    similarly expensive as training the models which
    constitute the ensemble.
    Nevertheless, as illustrated in Table~\ref{tab:distill},
    model soups still perform favorably.
    
    Table~\ref{tab:distill} also introduces stochastic augmentation.
    For each data point, stochastic augmentation randomly applies minimal, medium, or
    strong data augmentation.
    Additionally, Table~\ref{tab:fixaug} explores an alternative method
    for merging augmentations together.
    This augmentation policy, which we refer to as
    fix-aug, is introduced by \citet{fixaug}.
    For fix-aug, strong augmentation is used for all but
    the final epoch, which uses minimal augmentation.
    
    Figure~\ref{fig:swa} and Figure~\ref{fig:swa-in22k} apply model soups to solutions
    which already average along the fine-tuning trajectory.
    Methods for averaging along an individual optimization trajectory
    include exponential moving averages
    (EMA)~\cite{szegedy2016rethinking} and stochastic
    weight averages (SWA)~\cite{izmailov2018averaging}.
    We find that EMA and SWA can improve the accuracy of a single model but
    that model soups provide improvements even when applied to models which have weight-averaging along their trajectory.
    We try learning rates $10^{-5}$ and $3 \cdot 10^{-5}$ and three learning
    rate schedulers: constant, cosine annealing with restarts, and cosine annealing (all schedules have a short warm up period).
    In Figure~\ref{fig:swa} we fine-tune a CLIP pre-trained ViT-B/32, while Figure~\ref{fig:swa-in22k}
    fine-tunes an ImageNet-21k pre-trained ViT-B/32.
    
    Table~\ref{tab:sam} explores the relation between
    model soups and sharpness-aware minimization (SAM)~\cite{foret2021sharpnessaware}.
    In line with previous results, we find that SAM improves accuracy over vanilla fine-tuning. Souping two
    models trained with SAM improves over either individual model, although the magnitude of the gain is smaller than for vanilla fine-tuning. Souping models trained with and without SAM yields higher accuracy than souping models trained only with vanilla fine-tuning or only with SAM.
    
\begin{figure}
    \centering
    \includegraphics[width=0.5\textwidth]{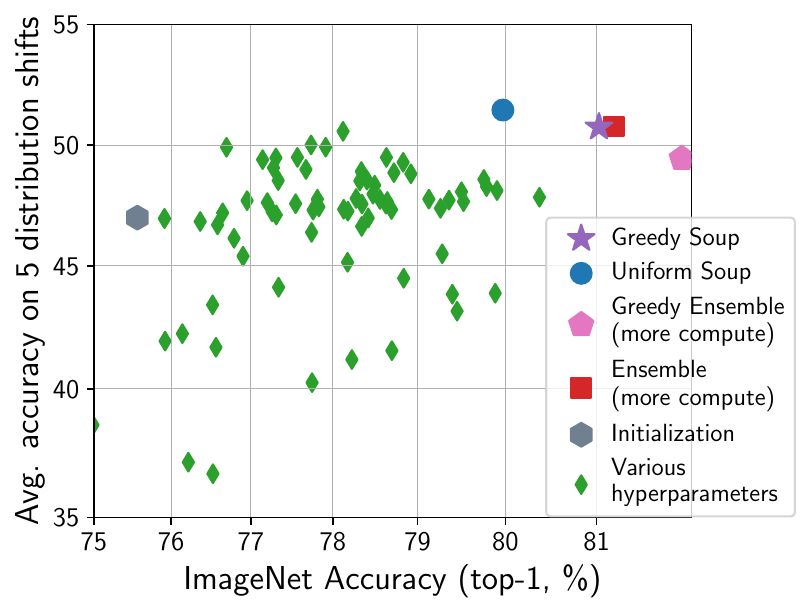}
    \caption{Adding additional results from Table~\ref{tab:results} to Figure~\ref{fig:teaser}.}
    \label{fig:teaser-ext}
\end{figure}

    As a final comparison that is potentially useful, we augment Figure~\ref{fig:teaser} with additional comparisons from Table~\ref{tab:results}. Results are shown in Figure~\ref{fig:teaser-ext}
    
    \begin{table}
        \caption{Comparing model soups to network distillation
        from an ensemble of models trained with different data augmentations.
        Stochastic data augmentation randomly applies minimal,
        medium, or strong data augmentation.}
        \begin{center}\footnotesize
        \setlength{\arrayrulewidth}{.01em}
        \begin{tabular}{lll}
            \toprule
            {} &              ImageNet &            Distribution shifts \\
            \midrule
            Individual model (LR 3e-05, minimal aug)                  &  76.42 &  43.21 \\
            Individual model (LR 3e-05, medium aug)                  &  78.83 &  43.55 \\
            Individual model (LR 3e-05, strong aug)               &  79.08 &  43.75 \\
            Individual model (LR 3e-05, stochastic aug)           &  78.94 &  45.04 \\
            Individual model (LR 3e-05, stochastic aug 3x epochs) &  78.38 &  42.18 \\
            Distillation from the ensemble (LR 3e-05, no aug)               &  78.59 &  43.45 \\
            Distillation from the ensemble (LR 3e-05, stochastic aug)       &  79.79 &  45.63 \\
            Soup minimal, medium, and strong aug (LR 3e-05)                              &  80.24 &  47.97 \\
            Ensemble minimal, medium, and strong aug (LR 3e-05)                          &  80.19 &  46.33 \\
            \midrule
            Individual model (LR 1e-05, minimal aug)                  &  77.19 &  47.98 \\
            Individual model (LR 1e-05, medium aug)                  &  79.51 &  46.74 \\
            Individual model (LR 1e-05, strong aug)               &  79.33 &  46.62 \\
            Individual model (LR 1e-05, stochastic aug)           &  79.48 &  48.07 \\
            Individual model (LR 1e-05, stochastic aug 3x epochs) &  79.59 &  46.89 \\
            Distillation from the ensemble (LR 1e-05, no aug)               &  79.13 &  47.28 \\
            Distillation from the ensemble (LR 1e-05, stochastic aug)       &  79.88 &  47.49 \\
            Soup minimal, medium, and strong aug (LR 1e-05)                              &  80.08 &  49.75 \\
            Ensemble minimal, medium, and strong aug (LR 1e-05)                          &  80.17 &  49.36 \\
            \bottomrule
            \end{tabular}
        \end{center}
        \label{tab:distill}
    \end{table}

    \begin{table}
        \caption{Comparing models soups of different augmentations with
        another method which combines different augmentation 
        strategies---fix aug, as described in \citet{fixaug}.
        For fix aug we use strong data augmentation for all
        except the final epoch for which we apply minimal aug.}
        \begin{center}\footnotesize
        \setlength{\arrayrulewidth}{.01em}
        \begin{tabular}{lll}
            \toprule
            {} &              ImageNet &            Distribution shifts \\
            \midrule
            Individual model (LR 3e-05, minimal aug)           &  76.42 &  43.21 \\
            Individual model (LR 3e-05,  medium aug)           &  78.83 &  43.55 \\
            Individual model (LR 3e-05, strong aug)        &  79.08 &  43.75 \\
            Individual model (LR 3e-05, fix aug)           &  79.43 &  45.46 \\
            Individual model (LR 3e-05, fix aug 4x epochs) &  78.57 &  41.53 \\
            Soup minimal, medium, and strong aug (LR 3e-05)             &  80.24 &  47.97 \\
            Soup minimal, medium, strong, and fix aug (LR 3e-05)         &  80.41 &  48.14 \\
            \midrule
            Individual model (LR 1e-05, minimal aug)           &  77.19 &  47.98 \\
            Individual model (LR 1e-05,  medium aug)           &  79.51 &  46.74 \\
            Individual model (LR 1e-05, strong aug)        &  79.33 &  46.62 \\
            Individual model (LR 1e-05, fix aug)           &  79.70 &  48.18 \\
            Individual model (LR 1e-05, fix aug 4x epochs) &  79.96 &  45.86 \\
            Soup minimal, medium, and strong aug (LR 1e-05)             &  80.08 &  49.75 \\
            Soup minimal, medium, strong, and fix aug (LR 1e-05)         &  80.17 &  49.71 \\
            \bottomrule
            \end{tabular}
        \end{center}
        \label{tab:fixaug}
    \end{table}

    \begin{table}
        \caption{Applying model soups to models trained with sharpness
        aware minimization (SAM)~\cite{foret2021sharpnessaware}.}
        \begin{center}\footnotesize
        \setlength{\arrayrulewidth}{.01em}
        \begin{tabular}{lll}
            \toprule
            {} &              ImageNet &            Distribution shifts \\
            \toprule
            Vanilla fine-tuning (seed 0)                    &  79.32 &  45.09 \\
            Vanilla fine-tuning (seed 1)                   &  79.16 &  45.12 \\
            SAM fine-tuning (seed 0) &  79.61 &  43.78 \\
            SAM fine-tuning (seed 1)                     &  79.59 &  43.79 \\
            \midrule
            Soup (vanilla fine-tuning, seeds 0 and 1)                 &  79.78 &  46.46 \\
            Soup (SAM fine-tuning, seeds 0 and 1)                   &  79.85 &  44.44 \\
            Soup (vanilla fine-tuning and SAM fine-tuning, seed 0)    &  80.04 &  45.38 \\
            \bottomrule
            \end{tabular}
        \end{center}
        \label{tab:sam}
    \end{table}

\end{document}